\documentclass[opre, nonblindrev]{informs3}

\OneAndAHalfSpacedXI 

\usepackage{endnotes}
\let\footnote=\endnote

\usepackage{booktabs}
\allowdisplaybreaks
\usepackage{dsfont, pifont, mathrsfs}
\usepackage{multirow}
\usepackage{setspace}
\usepackage[hidelinks, colorlinks=true, citecolor=blue]{hyperref}
\usepackage{algorithm} 
\usepackage{algorithmic}

\usepackage{natbib}
 \bibpunct[, ]{(}{)}{,}{a}{}{,}%
 \def\bibfont{\small}%

\newcommand{\w}{\mathbf{w}}

\TheoremsNumberedThrough     
\ECRepeatTheorems

\EquationsNumberedThrough    

\usepackage{macro}

\begin{document}

\RUNTITLE{Weak Shadow Variable for MNAR Data}

\TITLE{AI-Generated Measurements for Identification and Inference with Missing Data: A Weak Shadow Variable Approach}

\ARTICLEAUTHORS{%
\AUTHOR{Hongyu Chen\textsuperscript{1} \quad\quad David Simchi-Levi\textsuperscript{1} \quad\quad Ruoxuan Xiong\textsuperscript{2}}
\AFF{\textsuperscript{1}Massachusetts Institute of Technology, Cambridge, MA 02139\texorpdfstring{\\}{}%
\textsuperscript{2}Emory University, Atlanta, GA 30322\texorpdfstring{\\}{}%
\EMAIL{\texttt{chenhy@mit.edu, dslevi@mit.edu, ruoxuan.xiong@emory.edu}}}
} 

\ABSTRACT{%

Across business and social science applications, outcomes are often missing in ways that depend on the unobserved outcomes themselves. In service systems, for example, whether a customer submits a rating depends on the rating they would have provided. Such missing-not-at-random (MNAR) mechanisms make population quantities difficult to identify without strong assumptions on the observation process. Meanwhile, rich unstructured data, such as customer interaction histories, are increasingly available and can be used to construct structured measurements using tools such as large language models (LLMs). In this work, we develop an assumption-lean partial identification framework that uses such measurements as weak shadow variables, defined as outcome-informative proxies that are conditionally independent of missingness given the true outcome and observed covariates. Importantly, they need not accurately predict missing outcomes or satisfy the completeness requirement in the classical shadow variable literature. For identification, we characterize sharp bounds on population quantities through a pair of linear programs. For estimation and inference, we propose a localized penalized estimator that remains feasible under sampling error, and a subsampling algorithm for constructing confidence intervals. In semi-synthetic experiments using real customer-service dialogues, weak-shadow-variable intervals are about 89\% narrower than those without auxiliary information, while their midpoints have around 41\% lower estimation error than classical MNAR methods.

}%

\KEYWORDS{partial identification; missing not at random; shadow variables; large language models; linear programming}

\maketitle

\section{Introduction}

Missing data is pervasive in operations and social science research. In household and health surveys, respondents often skip questions perceived as sensitive or irrelevant. On digital platforms, users often choose whether or not to leave feedback based on their experiences. As noted by \citet{abrevaya2017gmm}, nearly 40\% of top economics papers report data missingness, with about 70\% dropping observations as a result.

In many of these settings, data is \textit{missing not at random} (MNAR), that is, the probability of observing an outcome depends on its possibly unobserved value. For example, \citet{bollinger2019trouble} shows that nonresponse across the earnings distribution is U-shaped, where left-tail ``strugglers'' and right-tail ``stars'' are least likely to report earnings. Conversely, an inverse U-shaped missingness pattern can be found in online product reviews, where users with more extreme opinions are more likely to leave reviews \citep{hu2017self}. This dependence between missingness and actual outcome creates a fundamental challenge for accurate estimation and decision-making because the observed data distribution may not be the same as the true data distribution. As a result, estimators based solely on observed data without modeling the missing mechanism can be severely biased, motivating approaches that explicitly account for the missingness mechanism.

In this paper, we study the problem of identifying and estimating population quantities, such as the mean outcomes, when outcomes are MNAR. Such questions are prevalent in service platforms or social surveys, e.g., when a platform seeks to evaluate average customer satisfaction, or when a researcher aims to estimate average income in a particular region. For those questions, one class of classical methods addresses this MNAR problem by imposing strong parametric structural assumptions, such as those in the Heckman selection model \citep{heckman1979sample} or the Pattern-Mixture model \citep{rubin1987calculation, little1994class}. Another common approach introduces auxiliary variables, including instrumental variables \citep{d2010new} or shadow variables \citep{miao2016varieties}, which need to satisfy restrictive exclusion or completeness conditions for identification. Both strategies can face practical limitations. In particular, structural parametric models may be misspecified, and identifying valid auxiliary variables may require substantial domain expertise or serendipity.

We therefore take a different approach and study, under realistically minimal assumptions, what we can still learn about population quantities like the mean outcomes. Instead of seeking point identification, we adopt a partial identification perspective \citep{manski2003partial} and characterize sharp upper and lower bounds on the estimand (e.g., mean outcome) when the distribution of the observed data can be known exactly. These partial identification bounds can be understood as the information limit of the observed data about the target outcome. Our key insight is that these bounds can be characterized by a pair of linear programs (LPs), whose objectives represent the estimand and whose constraints encode the probabilistic structure implied by the observed data. 

While bounds obtained under minimal assumptions are valid, they can be wide, especially when a large portion of the data is missing. To tighten the bounds, we use modern artificial intelligence (AI) systems, particularly large language models (LLMs), to extract information from rich, unstructured text. In many service settings, even when customer ratings are missing, platforms often retain interaction records such as chat transcripts. LLMs can convert these records into low-dimensional external measurements, including a predicted satisfaction score or an assessment of request resolution, that can be utilized for statistical inference. Recent work suggests that LLMs can approximate human behavior in complex settings \citep{horton2023large,goli2024frontiers,brand2024using}, making them promising tools for processing such context-rich records. Meanwhile, discrepancies between LLM outputs and actual human behavior have been documented \citep{gui2023challenge,li2024frontiers,gao2025take}, and researchers have cautioned against assuming that model predictions can perfectly substitute for human judgments. 

Given both the promise and limitations of these external evaluations, we propose a new structure to utilize the LLM-generated evaluations that both accounts for the MNAR pattern and the intrinsic bias in such evaluations. In particular, we treat LLM-generated outputs as \textit{weak shadow variables}, which are defined to be a fully observed auxiliary variable that satisfies an exclusion restriction from the missingness mechanism after conditioning on the outcome and covariates. This structure accommodates a wide range of special cases, for example, the evaluations can be random noise, perfect prediction, and predictions with fixed bias.

We use the term \textit{weak} because we do not impose the completeness condition required by classical shadow-variable methods \citep{miao2016varieties}.  Intuitively, completeness requires the auxiliary signal to distinguish outcome values, conditional on the covariates, well enough to uniquely recover the population quantity. This relaxation is particularly relevant for general-purpose AI models, which are typically pretrained on broad language tasks, and may therefore produce noisy or weakly informative measurements about the target outcome, violating the completeness condition.

\subsection{Main Contributions}

We summarize our four main contributions below.

First and foremost, we propose a realistic structure for utilizing external measurements under MNAR. Instead of assuming the measurements are accurate predictions of the missing outcomes, we only assume that they are independent of the missingness indicator once conditioned on the actual outcome and observed covariates. Such a weak shadow variable structure naturally incorporates perfect prediction as a special case and is easier to satisfy compared to other classical MNAR methods that use parametric models or strong auxiliary variables.  

Second, we propose a linear programming framework for partial identification both with and without weak shadow variables. In the baseline setting without auxiliary inputs, the formulation yields closed-form solutions for the identification interval of the mean outcome. When incorporating auxiliary predictions from LLMs, we derive analytical results that quantify how these predictions tighten the feasible set and narrow the identification interval. This formulation offers a unified and tractable approach to understand how predictive signals impact identification under minimal assumptions.

Third, we develop a finite-sample estimator for the endpoints of the identification interval and derive its asymptotic properties. Specifically, the estimator is based on a localized penalized linear program that accounts for estimation error in the probability constraints and remains feasible when their sample analogs are noisy or inconsistent. 
We establish consistency and convergence rates for the resulting estimator, showing that it achieves the same convergence rate as the first-stage probability estimators. We further develop a subsampling-based inference procedure for both the endpoints and the full identification region. This approach addresses the nonsmoothness and optimizer-switching behavior of the LP value function, where standard bootstrap procedures can fail. We establish consistency of the subsampling distribution and asymptotically valid confidence regions for the identified set.

Finally, we evaluate the proposed methods through simulation studies and semi-synthetic experiments using real customer-service dialogue data. To construct auxiliary signals, we generate outcome measurements using LLMs that target different aspects of the conversation from different perspectives. Our results reveal two key insights. First, we show that the resulting LLM measurements only add little or no prediction power to the rating probability after conditioning on the actual customer rating, supporting the weak shadow variable assumption in realistic settings. Second, we show that the measurements remain informative. In particular, incorporating LLM-based weak shadow variables reduces the width of identification intervals by 89\% across prompting strategies, and such improvement is not alleviated for simple binary shadow variables. 

\subsection{Related Work}

Our work relates to the growing literature on leveraging pretrained models as auxiliary signals to improve identification or statistical efficiency. One line of work is prediction-powered inference (PPI) \citep{angelopoulos2023prediction, angelopoulos2023ppi++, ji2025predictions}, which assumes that true labels are observed for only a random subset of the data, while predictions from an external model are available for all samples, with the aim of combining the two sources to enable valid inference. From a similar perspective, \citet{wang2024large} and \citet{lu2026generative} explore how LLM-generated simulations, when grounded in real data, can support accurate conjoint analysis in marketing research. \citet{wang2025efficient} further extends this framework by considering the fine-tuning of LLMs. Our work differs from these studies in that we allow the observed subsample of true labels to be selectively sampled, which is a more realistic but also significantly more challenging setting. Other literature along this line includes \citet{xu2026uncertainty}, which proposes using LLMs for semantic augmentation in causal inference, and \citet{li2025choosing}, which considers how to select a better algorithm under data sharing. Another stream of literature explores how to integrate imperfect AI-generated information into decision-making. \citet{ye2025lola} studies how to augment online learning algorithms with LLM-generated information. \citet{chen2025utilizing} examines how to design data-collection strategies in the presence of such LLM-based predictors. \citet{zhang2026llm} studies how to use transformers to share information for decision-making. These works assume the decision-maker can control the data generation process, meaning they can control whose ground-truth outcome to collect. Unlike these works, our work concerns a setting where the decision-maker cannot control the data collection process, and outcomes are MNAR. 

Methodologically, our work contributes to the rich literature on identification and estimation under MNAR mechanisms. Classical approaches include parametric selection models, such as the Heckman correction, which jointly models the outcome and missingness process \citep{heckman1979sample}, and Pattern-Mixture models that parameterize outcome distributions within each missingness stratum \citep{little1994class,rubin1987calculation}. Other strands of work leverage graphical models to represent missing data processes \citep{fay1986causal}, or use auxiliary variables such as instrumental variables that affect missingness but not outcomes \citep{das2003nonparametric,tchetgen2017general,sun2018semiparametric}. Our approach is most closely aligned with recent developments in the shadow variable literature \citep{d2010new,miao2016varieties,miao2024identification}, which typically uses the odds ratio to identify the distribution of missing outcomes. We contribute to this line of research in three key ways. First, we introduce a novel linear programming framework that characterizes the identification region for mean outcomes under MNAR. Second, we generalize the shadow variable approach by allowing weak shadow variables; this enables the use of auxiliary signals, e.g., from LLMs, that may violate classical completeness assumptions and thus do not yield point identification but can still significantly tighten bounds. Third, we establish convergence rates for our estimated identification region under both partial and point identification regimes.

Our linear programming formulation connects to the broader literature on inference for partially identified models \citep{manski2003partial, imbens2004confidence}. \citet{ChernozhukovHongTamer2007} proposed a criterion-function approach with set expansion to construct confidence regions for identified sets. \citet{beresteanu2008asymptotic} connect identified sets to LP optimal values through a support function characterization, and \citet{mogstad2018using} and \citet{kaido2019confidence} develop LP-based inference for treatment effect bounds and subvector projections, respectively. Our work derives a specific LP structure from the shadow variable assumption under MNAR and shows that auxiliary predictions generate additional constraints that tighten the identified set.

\smallskip
The remainder of the paper is organized as follows. Section~\ref{sec:setup} introduces the problem setup. Section~\ref{sec:lp} develops the linear programming framework for partial identification, both with and without weak shadow variables. Section~\ref{sec:set-expansion} presents the set-expansion estimator and its convergence properties. Section~\ref{sec:experiment} reports simulation and semi-synthetic experiments, and Section~\ref{sec:conclusion} concludes.


\section{Problem Setup}\label{sec:setup}


Suppose we are evaluating a social or digital system that solicits discrete feedback, such as program satisfaction ratings or customer-service reviews on online platforms. Outcomes are observed only when individuals choose to respond, and because not all individuals provide feedback, outcomes are only partially observed. We will use $R \in \{0,1\}$ to indicate whether a user's rating is observed ($R = 1$) or missing ($R = 0$). The actual rating is denoted by $Y$, and without loss of generality, we let the support of $Y$ be $[M]$, where $[M] = \{1, \ldots, M\}$ represents a discrete set of possible scores.

We focus on settings where missingness is not at random, meaning the probability of observing a rating may depend on its value, that is, $R \not\indep Y$. This phenomenon is common in practice. For example, users with extremely positive or extremely negative experiences are often more likely to leave reviews than users with moderate experiences \citep{hu2017self}. As a result, observed ratings may not be representative of the underlying population. We assume that the observed data form an i.i.d.\ sample $\{R_i, R_iY_i\}_{i=1}^n$, where $Y_i$ is observed only when $R_i=1$, and $n$ is the number of observations. In this section, we do not consider any observed covariates, which will be modeled in Section \ref{sec:lp}.

Our primary objective is to estimate the mean outcome
\[\theta = \E[Y],\]
such as the average rating across all customers. While we focus on the population mean for concreteness, our framework extends directly to other population quantities, including functionals of the form $\E[g(Y)]$ and other distributional summaries, as discussed later.

A central challenge is that, under missing-not-at-random sampling, the observed data generally do not contain enough information to uniquely determine $\theta$. For example, if customers with extreme experiences are more likely to provide feedback, the observed rating distribution may systematically differ from the rating distribution in the full population. Consequently, point identification of $\theta$ is impossible without introducing additional assumptions about the missingness mechanism, such as parametric selection models or exclusion restrictions based on auxiliary variables \citep{heckman1979sample, little1994class}. These assumptions are often difficult to verify empirically, and incorrect assumptions can lead to misleading point estimates.

Rather than imposing strong assumptions solely to obtain a single numerical estimate, we adopt a partial identification perspective, where the goal is to characterize the set of values of $\theta$ that are consistent with the observed data under minimal assumptions. We first introduce the baseline setting in which the observed data consist only of $\{R_i, R_iY_i\}_{i=1}^n$. Although the result in Section \ref{subsec:partial-identification-simplified} appears in \cite{manski2003partial}, but we provide a new perspective through an LP derivation, which will facilitate extending the result to the case with auxiliary information. In this sense, we think it is also beneficial for us to reiterate the result.

\subsection{Sharp Bounds without Auxiliary Information}\label{subsec:partial-identification-simplified}

Our partial identification strategy is based on the following decomposition of the population mean:
\[\theta = \sum_{y=1}^M y \cdot \P(Y = y) = \sum_{y=1}^M y \cdot \underbrace{\P(Y=y, R=1)}_{:=\alpha(y)} \big/ \underbrace{\P(R=1\mid Y=y)}_{:=\pi(y)} \,, \]
where the key is to express the marginal distribution $\P(Y=y)$ by the ratio between joint distribution on the observation data $\alpha(y):=\P(Y=y, R=1)$ and response probability $\pi(y):=\P(R=1\mid Y=y)$. If we are interested in other population quantities (e.g., $\E[g(Y)]$), then we replace $y \cdot \P(Y = y)$ by $g(y) \cdot \P(Y = y)$ in the decomposition.
Here, the joint probability $\alpha(y)$ is identifiable from observed data, i.e., after observing the data $\{R_i, Y_iR_i\}_{i=1}^n$, we are able to estimate $\alpha(y)$ consistently as it only corresponds to the observation sample ($R=1$). However, the conditional response probability $\pi(y)$ is generally unidentifiable when missingness depends on the outcome itself because we cannot observe the missing outcomes. As a result, the mean $\theta$ cannot be point-identified without further assumptions.

Although we cannot recover the mean exactly from observation data, we can characterize the identification region that $\theta$ can take values in. Since $\pi(y)$ is the only parameter that cannot by identified from data, we proceed this goal by considering all possible values of $\pi(y) \in (0,1]$. Note that the only constraint from observational data is that the probabilities $\P(Y=y)$ induced by the response probability $\pi(y)$ must sum to one. Thus, the feasible set for $\pi(y)$ can be written as
\[ \Pi= \left\{ (\pi(1), \dots, \pi(M)): \sum_{y=1}^M \alpha(y) / \pi(y)  = 1, \pi(y)\in(0,1]\right\} \,, \]
where the constraint is exactly that $\sum_y\P(Y=y)=1$. Each of the point in $\Pi$ defines a possible underlying distribution that is compatible with the observation distribution. From only the observation data, we cannot distinguish any two compatible distributions. By calculating the mean outcome for every compatible distribution induced by $\Pi$, we can have the sharp identification set for $\theta$, i.e.,
\[\Theta =\left\{ \sum_{y=1}^M y \cdot \alpha(y)/\pi(y): (\pi(y))_{y\in[M]}\in\Pi\right \},\]
where $\sum_{y}y \cdot \alpha(y)/\pi(y)=\sum_yy\P(Y=y)=\E[Y]$ represents the target parameter.

To simplify notation, let $w(y)=1/\pi(y)-1$. Under this change of variables, the feasible region $\Pi$ becomes a polyhedron in $w(y)$, and the mapping from $w(y)$ to $\theta$ is linear. Thus, the identification region $\Theta$ is a closed interval, and its endpoints can be computed by solving the following pair of linear programs (LPs):
\begin{equation}\label{opt:mean-mnar-range}
\begin{aligned}
\begin{minipage}{0.45\textwidth}
\[
\begin{aligned}
     \theta_{\min} = \min_{w(y)} & \quad \sum_{y=1}^M y \cdot \alpha(y) (w(y)+1) \\
     \text{s.t.} & \quad \sum_{y=1}^M \alpha(y) (w(y)+1) = 1 \\
     & \quad w(y)\geq 0 \quad \forall y
\end{aligned}
\]
\end{minipage}
\quad
\begin{minipage}{0.45\textwidth}
\[
\begin{aligned}
     \theta_{\max} = \max_{w(y)} & \quad \sum_{y=1}^M y \cdot \alpha(y) (w(y)+1) \\
     \text{s.t.} & \quad \sum_{y=1}^M \alpha(y)(w(y)+1) = 1 \\
     & \quad w(y)\geq 0 \quad \forall y
\end{aligned}
\]
\end{minipage}
\end{aligned}
\end{equation}

Therefore, the identified region for the mean outcome is given by $\Theta = [\theta_{\min}, \theta_{\max}]$. In the proposition below, we show that we can obtain both $\theta_{\min}$ and $\theta_{\max}$ analytically.

\begin{proposition}\label{prop:no-shadow-bound}
The sharp identification region for $\theta$ given observed MNAR data $\{R_i, R_iY_i\}_{i=1}^n$ is $\Theta = [\theta_{\min}, \theta_{\max}]$ defined in \eqref{opt:mean-mnar-range}, whose endpoints have the following closed-form expressions:
    \begin{align*}
        \theta_{\min} =&~ \P(R=1)\E[Y\mid R=1] + \P(R = 0)\\
        \theta_{\max} =&~ \P(R=1)\E[Y\mid R=1] + M \cdot \P(R = 0).
    \end{align*}
\end{proposition}

Here we implicitly let $\P(R=1)\E[Y\mid R=1]$ be zero if $\P(R=1)$ is zero and hence $\E[Y\mid R=1]$ is undefined. These expressions are attained by setting the weights $w(y)$ to their minimum allowable value $w(y) = 0$ for all but one outcome level. To achieve $\theta_{\min}$, we set $w(y) = 0$ for $y = 2, \ldots, M$ and assign the remaining mass to $y = 1$, the smallest outcome. Conversely, to achieve $\theta_{\max}$, we set $w(y) = 0$ for $y = 1, \ldots, M-1$ and concentrate the remaining weight on $y = M$, the largest outcome. This corresponds to placing as much probability mass as possible on the lowest or highest feasible rating levels, subject to the constraint induced by the observed distribution $\alpha(y) = \P(Y = y, R = 1)$. 

Notably, the width of the identification region is $\theta_{\max} - \theta_{\min} = (M-1) \cdot \P(R = 0)$, which scales linearly with the probability of missingness. When $\P(R = 0) = 0$, i.e., outcomes are fully observed, the bounds collapse to a point and $\theta$ is point-identified. In contrast, when $\P(R = 0) = 1$, the bounds are equal to the full support range, $[1, M]$, which is uninformative. Without additional information and structural assumptions, the identification region in Proposition~\ref{prop:no-shadow-bound} is the best one can hope for. These bounds are sharp in the sense that any other valid identification region from observed data will contain $\Theta$ as a subset.

\section{Partial Identification with Weak Shadow Variables}\label{sec:lp} \label{subsec:partial-identification-with-shadow}
In many modern applications, a missing outcome does not mean that all information about the unit is missing. Even when the final rating $Y$ is unobserved, the platform often records rich contextual information $W$, such as dialogue transcripts or service logs. Because this information is high-dimensional and unstructured, it has been difficult to incorporate into classical missing-data analyses. Modern LLMs make this information more accessible by transforming text and interaction records into structured measurements of service quality, such as a numerical rating. In this section, we study how to use this structured external measurement to tighten the identification region in Equation \eqref{opt:mean-mnar-range}.

The unstructured text data $W$ usually contains several distinct types of information. To understand the distinct roles of information in $W$, we first let $X_0$ denote the observed structured covariates, e.g., customer type, age, gender, etc., and conceptually decompose the information in $W$ into three parts $W=(W_R, W_Y, W_0)$. The component $W_R$ contains information that may directly affect the response decision $R$ and may also be related to outcome $Y$ (e.g., customer's personality). The component $W_Y$ is designed to be more informative about $Y$ (e.g., task completion quality), but is excluded from the response mechanism once $Y$, $X_0$, and $W_R$ are held fixed: $W_Y \indep R \mid Y, X_0, W_R$.  Finally, $W_0$ contains residual information relevant to neither $R$ nor $Y$: $W_0 \indep (R, Y) \mid X_0, W_R, W_Y$. 

We then use LLM prompts to extract distinct types of information in $W$. In particular, we construct outcome-focused measurements $F$ that intend to capture information in $W_Y$, such as service quality, answer correctness, or task completion. In parallel, we can construct response-focused summaries of $W_R$ such as customer types and combine them with $X_0$ to form the covariate vector $X$. We provide practical guidance for constructing these summaries in Section \ref{subsec:guidance}. Because $F$ is intended to preserve the inferential role of $W_Y$, it is then natural to impose an exclusion restriction on $F$, formalized in the following assumption.

\begin{assumption}[Weak Shadow Variable Exclusion]\label{ass:cond-indep}
The outcome-focused measurements $F$ are excluded from the missingness mechanism given the true outcome $Y$ and covariates $X$. Equivalently,
\[
    F \indep R \mid Y,X.
\]
\end{assumption}

We refer to any variable $F$ satisfying Assumption~\ref{ass:cond-indep} as a \emph{Weak Shadow Variable}. The central feature of this assumption is that we are able to condition on the (possibly missing) true outcome $Y$, which provides rich information about both the measurement $F$ and missingness $R$, thereby allowing this assumption to accommodate a wide range of special cases. For example, an independently generated white noise satisfies the assumption, although it provides no additional identifying information. At the other extreme, the assumption holds automatically when $F$ is a deterministic function of $(Y,X)$, including the perfect prediction case $F=Y$, or the prediction with a fixed group-specific bias case $F=Y+f(X)$ for an arbitrary function $f$. As another mental model, Assumption~\ref{ass:cond-indep} holds whenever the response mechanism takes the form $\Pr(R=1\mid F,Y,X)=\phi(Y,X)$ for some unknown function $\phi$, which is a common model and can be seen in literature such as \cite{tang2003analysis} and \cite{dellarocas2008sound}. 

This exclusion restriction also appears in classical shadow-variable methods. Our framework generalizes those methods by imposing neither a minimum relevance requirement nor the completeness condition typically used for point identification (see Remark 1), which is why we call $F$ a \textit{weak} shadow variable. Similar to other identification assumptions, Assumption~\ref{ass:cond-indep} is not directly testable from observed data. To assess robustness to departures from this assumption, Section~\ref{subsec:shadow-sensitivity} develops a sensitivity analysis that quantifies how violations of the exclusion restriction affect the identified set.



\begin{remark}[Connection to Shadow Variables]
    Classical shadow-variable methods require an additional \emph{completeness condition} to point-identify the full-data distribution \citep{d2010new,miao2016varieties,miao2024identification}. The conditional distribution $\P(Y\mid X, F, R=1)$ is complete if, for every square-integrable function $h(X,Y)$, $\E[h(X,Y)\mid X, F, R=1]=0$ almost surely implies $h(X,Y)\equiv0$ almost surely. Intuitively, completeness requires variation in $F$ to distinguish the possible outcome values sufficiently well to uniquely recover the missing-outcome distribution. This condition can be restrictive; for example, a binary $F$ generally cannot satisfy completeness for a nonbinary $Y$. When completeness holds, the identification interval collapses to a point. When it fails, our framework instead characterizes the resulting partial-identification region. See Appendix \ref{apx.sec:connection} for more discussion.
\end{remark}

\begin{remark}[Comparison with MAR assumption]
    The classical Missing-at-Random assumption requires $Y\indep R\mid X$ for a sufficiently rich set of covariates $X$ \citep{rubin1976inference, robins1994estimation, qin2008efficient}. Thus, it requires the stability of outcome $Y$ across observation and missing samples for a strata depends on $X$. On the other hand, the shadow variable structure represents the stability of the measurement $F$ on a finer strata that is determined by both the outcome $Y$ and covariates $X$. In applications involving post-outcome evaluations, the shadow variable structure is more plausible as same technological tool is used for all the interaction, making it stable across the observation and control group. See \cite{rambachan2024program} and \cite{dell2026measurement} for more discussion.
\end{remark}

\subsection{Identification Interval}
In this section, we develop the sharp identification region for $\theta$ with weak shadow variables. In this setting, the observed data form an i.i.d. sample of $\{X_i, F_i, R_i, R_iY_i\}_{i=1}^N$. The high-level idea is the same as in Section \ref{subsec:partial-identification-simplified}, i.e., we first specify all the possible data distributions that are compatible with the observations, and then calculate the average outcome on that compatible set. Here, our partial identification strategy is based on the decomposition $\theta = \mathbb{E}[\theta_X]$, where $\theta_x = \mathbb{E}[Y \mid X = x]$ is the conditional mean given covariate value $X = x$. We will first provide an identification region for every $x \in \mathcal{X}$ and then aggregate these regions to obtain an identification region for $\theta$. For each $x \in \mathcal{X}$, we have the following decomposition
\begin{align*}
    \theta_x =&~ \sum_{f \in \mathcal{F}}  \sum_{y = 1}^M y \cdot \P(F=f, Y=y\mid X=x) \\
    =&~ \sum_{f \in \mathcal{F}}  \sum_{y = 1}^M y \cdot {\P(R=1, F=f, Y=y \mid X=x)}\big/{\P(R=1\mid F=f, Y=y, X=x)}\,.\\
    =&~ \sum_{f \in \mathcal{F}}  \sum_{y = 1}^M y \cdot \underbrace{\P(R=1, F=f, Y=y \mid X=x)}_{:=\alpha_x(f,y)}\big/\underbrace{\P(R=1\mid Y=y, X=x)}_{:=\pi_x(y)}\,.
\end{align*}
The second equality follows from the chain rule of probability. Under Assumption \ref{ass:cond-indep}, i.e., $F\indep R\mid Y,X$, we have $ \P(R = 1 \mid Y = y, X = x) = \P(R = 1 \mid F = f, Y = y, X = x)$ for all $f \in \mathcal{F}$, so the denominator does not depend on $f$, and we write it as $\pi_x(y)$ for notational simplicity.

The quantity $\alpha_x(f, y)$ is identifiable from observed data, but $\pi_x(y)$ remains unidentifiable. We therefore propose to identify a set of feasible values for $\pi_x(y)$ by leveraging the following identity that connects the marginal distribution of the shadow variable in the missing group to the observation group,
\begin{equation*}
    \begin{aligned}
        \underbrace{\P(R=0, F=f\mid X=x)}_{:=\beta_x(f)}&=\sum_{y=1}^M\P(R=0, F=f, Y=y\mid X = x) \\
        &=\sum_{y=1}^M\P(F=f, Y=y \mid X=x)\P(R=0\mid Y=y, X=x)\\
        &=\sum_{y=1}^M\frac{\alpha_x(f, y)}{\pi_x(y)} \cdot (1 - \pi_x(y))\,,
    \end{aligned}
\end{equation*}
where the second equality uses Assumption \ref{ass:cond-indep}. We let $\beta_x(f)=\P(R=0, F=f\mid X=x)$ for notational simplicity. Note that $\beta_x(f)$ is the distribution of the shadow variable in the missing population, which is identifiable from observed data. Thus, the only free parameter is $\pi_x(y)$ in this case, and for each $x \in \mathcal{X}$, we can similarly write the feasible region for $\pi_x(y)$ as
\[ \Pi_x= \left\{ (\pi_x(1), \dots, \pi_x(M)): \sum_{y=1}^M\alpha_x(f,y)\left(\frac{1}{\pi_x(y)}-1\right)=\beta_x(f)\,,~\;\forall f \in \mathcal{F},  \pi_x(y)\in(0,1] \right\} \]
and the identification region for $\theta_x$ becomes
\[ \Theta_x=\left\{ \sum_{f \in \mathcal{F}} \sum_{y=1}^M y \frac{\alpha_x(f,y)}{\pi_x(y)}: (\pi_x(y))_{y\in[M]}\in\Pi_x \right\} \,. \]


Letting \textbf{$w_x(y)=1/\pi_x(y)-1$}, we obtain a linear representation of the objective and constraints in terms of $w_x(y)$. The feasible set $\Pi_x$ remains convex, and so the identification region $\Theta_x$ is a closed interval. The endpoints are obtained by solving the following pair of linear programs:
%
\begin{equation}\label{opt:mnar-with-shadow-matrix-form-theta}
\begin{aligned}
\begin{minipage}{0.45\textwidth}
\[
\begin{aligned}
\theta_{x,\min} = \min_{\mathbf{w}_x} \quad & \bm{1}^\top A_x D(\mathbf{w}_x + \bm{1}) \\
\text{s.t.} \quad & A_x \mathbf{w}_x = \bm{\beta}_x \\
& \mathbf{w}_x \geq \bm{0}
\end{aligned}
\]
\end{minipage}
\quad
\begin{minipage}{0.45\textwidth}
\[
\begin{aligned}
\theta_{x,\max} = \max_{\mathbf{w}_x} \quad & \bm{1}^\top A_x D(\mathbf{w}_x + \bm{1}) \\
\text{s.t.} \quad & A_x \mathbf{w}_x = \bm{\beta}_x \\
& \mathbf{w}_x \geq \bm{0}
\end{aligned}
\]
\end{minipage}
\end{aligned}
\end{equation}
where $A_x=[\alpha_x(f,y)]_{f,y}\in [0,1]^{|\mathcal{F}|\times M}$, $\mathbf{w}_x=(w_x(1), \dots, w_x(M))^\top$, $\bm{\beta}_x=(\beta_x(1), \dots,\beta_x(|\mathcal{F}|))^\top$, and $D=\mbox{diag}\{1,2,\ldots,M\}$. The vectors $\bm{1}$ and $\bm{0}$ contain all ones and all zeros, respectively. The constraints restrict $\pi_x(y)$ to lie in $\Pi_x$, and the objective defines $\theta\in\Theta_x$. Aggregating over the covariate distribution $P_X$, we obtain the identification region for $\theta$.

\begin{theorem}\label{thm:point-id-mean}
    Under Assumption \ref{ass:cond-indep}, the sharp identification region for $\theta$ is given by \[\Theta = \left[\mathbb{E}_X\big[\theta_{X,\min}\big],~\mathbb{E}_X\big[\theta_{X,\max}\big] \right]:=\left[\theta_{\min}^{\mathrm{shad}},~ \theta_{\max}^{\mathrm{shad}}\right] \,, \]
    where $\theta_x$ is defined in \eqref{opt:mnar-with-shadow-matrix-form-theta}. Moreover, $\theta$ is point identified if $A_x$ has full column rank for all $x \in \mathcal{X}$.
\end{theorem}

The identification of $\theta$ depends on the identification of the conditional mean $\theta_x$, which is governed by the linear system $A_x$. If $A_x$ has full column rank, i.e., $\operatorname{rank}(A_x)=M$, then the linear constraint system has a unique solution, and we achieve point identification: $\theta_{x,\min} = \theta_{x,\max}$. This rank condition corresponds to the completeness condition for a classical shadow variable. More generally, if some rows of $A_x$ are linearly dependent (i.e., $F \indep Y \mid X, R = 1$ for every $F \in \mathcal{F}'$ for some subset $\mathcal{F}'\subset\mathcal{F}$) or if the columns are dependent (i.e., $F \indep Y \mid X, R = 1$ for a subset of outcomes $Y \in \mathcal{Y}' \subset \{1, \ldots, M\}$), then the feasible region contains multiple solutions and $\mathbf{w}_x$ is only partially identified. In this sense, our formulation generalizes the classical shadow variable approach: it allows violations of the completeness condition of the shadow variable definition and quantitatively characterizes how the strength of association between $F$ and $Y$ impacts the width of the identification region $\Theta_x$. Again, the identification in Theorem~\ref{thm:point-id-mean} is sharp in the sense that any other valid identification region based on the observed data will contain $\left[\theta_{\min}^{\mathrm{shad}}, \theta_{\max}^{\mathrm{shad}}\right]$ as a subset.

Lastly, to understand the effect of the additional prediction $F$, we compare the above identification region with the one defined in Equation \eqref{opt:mean-mnar-range}, where the shadow variable is not available. The formulation in linear program \eqref{opt:mnar-with-shadow-matrix-form-theta} is closely related to that in linear program \eqref{opt:mean-mnar-range}, whose lower-bound constraint can be written as the single aggregated constraint $\bm{1}^\top A_x \mathbf{w}_x = \bm{1}^\top \bm{\beta}_x $. Thus, we can use the techniques in aggregation bounds \citep{zipkin1980bounds, litvinchev2013aggregation} to analyze their differences.

\begin{proposition}\label{prop:lp-bound}
Write the matrix $A_x=[\bm{a}_{x,1}, \bm{a}_{x,2}, \dots, \bm{a}_{x, M}]$ for $P_X$-almost every $x$. On covariate values with $\bm{1}^\top\bm{\beta}_x>0$, assume the extreme observed columns have positive sums when they enter the normalized terms below; covariate values with zero denominators are interpreted as contributing zero to the corresponding lower bound. Then
\[\min\{\theta_{\max}-\theta_{\max}^{\mathrm{shad}},\; \theta_{\min}^{\mathrm{shad}}-\theta_{\min}\} \geq \mathbb{E}_X\left[\frac{\bm{1}^\top\bm{\beta}_X}{2}\left\|\frac{\bm{\beta}_X}{\bm{1}^\top\bm{\beta}_X}-\frac{\bm{a}_{X,M}}{\bm{1}^\top\bm{a}_{X,M}}\right\|_1 \right]\geq0 \,,\]
\end{proposition}
Proposition~\ref{prop:lp-bound} characterizes the difference between the identification bounds with and without a shadow variable. The proof is provided in Appendix \ref{apx:prop-lp-bound-proof}. As a special case, the identification region with a shadow variable is never wider than that without one, i.e., $\theta_{\min}\leq\theta_{\min}^{\mathrm{shad}}\leq\theta_{\max}^{\mathrm{shad}}\leq\theta_{\max} $. Moreover, the amount of improvement depends on the missingness ratio, represented by $\bm{1}^\top\bm{\beta}_X$, and the misalignment between the missing distribution and the observed extreme-outcome distribution, as captured by the $\ell_1$ terms inside the expectation. Thus, the shadow variable is especially useful when the fraction of missingness is high and when the missingness scheme does not align well with the observed scheme, in which case the data are far from MAR.

\begin{remark}[Prediction Bias and Informativeness]
    In our framework, $F$ enters only as an auxiliary measurement through the constraints summarized by $A_x$ and $\bm\beta_x$. Unlike methods that directly impute $Y$ with $F$ \citep{angelopoulos2023prediction}, our approach does not require either calibration or unbiasedness. Nevertheless, the informativeness of $F$ affects the interval width: noisy or coarse measurements may yield a rank-deficient $A_x$ and wide intervals, whereas more informative measurements tighten the bounds; full column rank for every $x$ yields point identification.
\end{remark}

\begin{remark}[Multiple Shadow Variables]
    In practice, multiple pretrained models may produce distinct predictive signals, which our framework naturally accommodates by jointly imposing their corresponding restrictions. Specifically, let $F^{(k)}$, for $k=1,\ldots,K$, denote the output of the $k$-th pretrained model. For each $k$ and covariate value $x$, let $A_x^{(k)}$ and $\bm\beta_x^{(k)}$ denote the corresponding matrix and vector defined as in \eqref{opt:mnar-with-shadow-matrix-form-theta}, and define $\mathcal W_x^{(k)}
    :=
    \bigl\{\mathbf w_x\ge\bm0:A_x^{(k)}\mathbf w_x=\bm\beta_x^{(k)}\bigr\}$ as the feasible region for the $k$-th shadow variable. Then the aggregated feasible region can be defined as
    \[
    \mathcal W_x^{\mathrm{multi}}
    :=
    \bigcap_{k=1}^K\mathcal W_x^{(k)}
    =
    \bigl\{\mathbf w_x\ge\bm0:
    A_x^{(k)}\mathbf w_x=\bm\beta_x^{(k)}\ \text{for all }k\in[K]\bigr\}.
    \]
    Let $\bar{\bm\alpha}_x=(\P(R=1,Y=1\mid X=x),\ldots,\P(R=1,Y=M\mid X=x))^\top$, so that $\bar{\bm\alpha}_x^\top=\bm1^\top A_x^{(k)}$ for every $k$. The conditional bounds based on all $K$ shadow variables are then obtained by optimizing over the common feasible region:
    \[
    \theta_{x,\min}^{\mathrm{multi}}
    =
    \min_{\mathbf w_x\in\mathcal W_x^{\mathrm{multi}}}
    \bar{\bm\alpha}_x^\top D(\mathbf w_x+\bm1),
    \qquad
    \theta_{x,\max}^{\mathrm{multi}}
    =
    \max_{\mathbf w_x\in\mathcal W_x^{\mathrm{multi}}}
    \bar{\bm\alpha}_x^\top D(\mathbf w_x+\bm1).
    \]
    Aggregating over $X$ yields
    \[
    \Theta^{\mathrm{multi}}
    =
    \left[
    \E_X\bigl[\theta_{X,\min}^{\mathrm{multi}}\bigr],
    \E_X\bigl[\theta_{X,\max}^{\mathrm{multi}}\bigr]
    \right].
    \]
    At the population level, the true odds vector belongs to $\mathcal W_x^{\mathrm{multi}}$ whenever every $F^{(k)}$ satisfies Assumption~\ref{ass:cond-indep}. Moreover, because $\mathcal W_x^{\mathrm{multi}}\subseteq\mathcal W_x^{(k)}$ for every $k$, the resulting interval is no wider than any interval obtained from a single shadow variable. 
\end{remark}

\subsection{Sensitivity Analysis}\label{subsec:shadow-sensitivity}
In this section, we study the robustness of our identification results to violations of Assumption~\ref{ass:cond-indep}. Our analysis relies on excluding the outcome-focused summary $F$ from the response mechanism given $(Y,X)$. In practice, however, this exclusion restriction may hold only approximately. For example, a model-generated assessment may capture aspects of the interaction that are correlated with both the user's latent rating and the user's willingness to provide feedback. In such cases, $F$ may contain residual information about $R$ beyond $(Y,X)$.

To quantify such departures, we introduce a sensitivity parameter that measures the extent to which the response probability can vary with $F$ after conditioning on $(Y,X)$. For a fixed covariate value $x$, let $\pi_x(f,y)=\P(R=1\mid F=f,Y=y,X=x)$ denote the response probability, let $\bar\pi_x(y)=\P(R=1\mid Y=y,X=x)$ denote the aggregated response probability, and define $w_x(y)={(1-\bar\pi_x(y))}/{\bar\pi_x(y)}$ and
$u_x(f,y)={(1-\pi_x(f,y))}/{\pi_x(f,y)}$ as the corresponding odds ratios.

Then, under Assumption~\ref{ass:cond-indep}, these two response probabilities coincide for every $f$, and hence $u_x(f,y)=w_x(y)$. When Assumption~\ref{ass:cond-indep} is no longer satisfied, we define a sensitivity radius that quantifies such violation.
\begin{assumption}\label{ass:shadow-sensitivity}
There exists a known constant $\rho\ge 0$ such that for every $x\in\mathcal X$, $f\in\mathcal F$, and $y\in[M]$,
\[
e^{-\rho} w_x(y)\le u_x(f,y)\le e^\rho w_x(y).
\]
\end{assumption}

Whenever $\pi_x(f,y),\bar\pi_x(y)\in(0,1)$, Assumption \ref{ass:shadow-sensitivity} is equivalent to requiring the two logits to differ by at most $\rho$. When $\rho=0$, Assumption~\ref{ass:shadow-sensitivity} reduces to Assumption~\ref{ass:cond-indep}. The odds-ratio formulation is useful because it preserves linearity in the feasible region. In particular, it introduces the auxiliary variables $u_x(f,y)$ to describe violations of the exclusion restriction, while the estimand remains a linear functional of the marginal odds $w_x(y)$. For each $x\in\mathcal X$, define $\theta_{x,\min}^{(\rho)}$ as the optimal value of
\begin{equation}\label{opt:shadow-sensitivity-lp}
\begin{aligned}
\theta_{x,\min}^{(\rho)}=\min_{w_x(\cdot),\,u_x(\cdot,\cdot)} \quad
& \sum_{y=1}^M y\,\alpha_x(y)\{1+w_x(y)\} \\
\text{s.t.}\quad
& \sum_{y=1}^M \alpha_x(f,y)\,u_x(f,y)=\beta_x(f), && \forall f\in\mathcal F,\\
& \sum_{f\in\mathcal F}\alpha_x(f,y)\,u_x(f,y)=\alpha_x(y)\,w_x(y), && \forall y\in[M],\\
& e^{-\rho}w_x(y)\le u_x(f,y)\le e^\rho w_x(y), && \forall (f,y)\in\mathcal F\times[M],\\
& w_x(y)\ge 0,\quad u_x(f,y)\ge 0, && \forall (f,y)\in\mathcal F\times[M].
\end{aligned}
\end{equation}
The upper endpoint $\theta_{x,\max}^{(\rho)}$ is obtained by replacing $\min$ by $\max$ in \eqref{opt:shadow-sensitivity-lp}.

\begin{theorem}\label{thm:shadow-sensitivity-lp}
Under Assumption~\ref{ass:shadow-sensitivity}, the interval $\Theta^{(\rho)}=[\theta_{\min}^{(\rho)},\theta_{\max}^{(\rho)}]:=[\mathbb{E}_X[\theta_{X,\min}^{(\rho)}], \mathbb{E}_X[\theta_{X,\max}^{(\rho)}]]$ is the sharp identification region for $\theta$, where $\theta_{\min}^{(\rho)}$ and $\theta_{\max}^{(\rho)}$ are defined in \eqref{opt:shadow-sensitivity-lp}.  When $\rho=0$, \eqref{opt:shadow-sensitivity-lp} reduces to the baseline weak-shadow-variable LP in \eqref{opt:mnar-with-shadow-matrix-form-theta}.
\end{theorem}

Theorem~\ref{thm:shadow-sensitivity-lp} shows that we can use a similar linear program to conduct sensitivity analysis. The only difference is that by allowing the weaker Assumption \ref{ass:shadow-sensitivity}, we transform the original equality constraint into an inequality constraint. Moreover, the sensitivity radius has a transparent monotonic effect, that is, if $0\le \rho_1\le \rho_2$, then every feasible point under $\rho_1$ is feasible under $\rho_2$, so $\Theta^{(0)}\subseteq \Theta^{(\rho_1)}\subseteq \Theta^{(\rho_2)}$. This provides a direct robustness curve for the identified set as the allowed deviation from exclusion increases.

\subsection{Practical Guidance}\label{subsec:guidance}
This section provides practical guidance for constructing candidate weak shadow variables $F$ from contextual information $W$ using LLM prompts. Different prompts target different aspects of $W$ and therefore produce signals with different identifying properties. Constructing $F$ involves balancing two objectives: it should be informative about the outcome $Y$ (\textsc{relevance}) and provide no additional information about the response decision $R$ once $(Y, X)$ are held fixed (\textsc{exclusion}). Improving \textsc{relevance} may weaken \textsc{exclusion}, creating a tradeoff between the two objectives. 

First, prompts should target outcome-relevant features of the underlying interaction. In customer-service applications, natural targets include service quality, answer correctness, task completion, and successful resolution. These features are likely to be informative about $Y$, while their association with $R$ is plausibly mediated by $Y$ and $X$. 

Second, prompts should avoid response-relevant or highly user-specific information. Incorporating interaction histories, demographics, behavioral patterns, or detailed persona information may improve outcome prediction but also increase the likelihood that $F$ captures factors directly related to response propensity, thereby weakening exclusion. When possible, such information should instead be incorporated into the covariates $X$.

In Section~\ref{sec:experiment}, we use semi-synthetic customer-service dialogue experiments to assess candidate shadow variables along both dimensions. The results document a tradeoff between \textsc{relevance} and \textsc{exclusion}, while indicating that departures from Assumption~\ref{ass:cond-indep} are relatively limited.

\section{Estimation and Inference in Finite Samples}\label{sec:set-expansion}

The identification results in Section~\ref{sec:lp} characterize the sharp identified set of $\theta$ in terms of the unknown quantities $A_x$ and $\bm\beta_x$, which need to be estimated from data. In this section, we study estimation and inference for the endpoints $\theta_{\min}$ and $\theta_{\max}$ of this set, explicitly accounting for the first-stage estimation error in $A_x$ and $\bm\beta_x$. These endpoints have a direct policy interpretation in the sense that they represent the most pessimistic and most optimistic population means that remain consistent with the observed data and the weak-shadow-variable assumption. Inference for the full identified set can then be obtained by combining confidence bounds for the two endpoints, as shown in Proposition~\ref{prop:ci-identification-region}.



A natural approach is to estimate $A_x$ and $\bm\beta_x$ from the data, plug these estimates into the linear program in~\eqref{opt:mnar-with-shadow-matrix-form-theta}, and solve the resulting sample linear program. Although simple, this direct plug-in approach is not reliable in finite samples for two reasons. First, the sample linear program may be infeasible. The population constraints are exact probability identities, but their empirical counterparts are subject to sampling error. When the linear system contains redundant or nearly redundant constraints, even small estimation errors can make the sample constraints mutually inconsistent. Second, even when the sample problem is feasible, the optimal value of a linear program can be sensitive to small changes in the estimated coefficients. A small perturbation of the empirical probabilities can change which constraints bind or which extreme point is optimal. This lack of smoothness makes standard plug-in inference and ordinary bootstrap procedures unreliable \citep{fang2019inference, goff2024inference}.

To address these two issues, we propose a localized penalized linear program in Section \ref{subsec:finite-sample-estimator}. The construction has two components. First, instead of imposing the equality constraint exactly, we allow violations of the constraint but penalize the violations in the objective. This softens the sample constraints and ensures that the optimization problem is always feasible. Second, we restrict the decision variables to a bounded box. This localization prevents unstable solutions and makes the value of the optimization problem Lipschitz continuous in the estimated probabilities. Importantly, this modification does not change the population target once the box is large enough to contain an optimal solution of the original linear program.



For notational simplicity, we suppress the stratum index $x$ and work conditionally on a fixed stratum. The same construction applies to discrete covariates by stacking the constraints across strata. Extensions to continuous covariates are discussed in Section~\ref{subsec:continuous-covariates}.

Let $O_i=(R_i,R_iY_i,F_i)$, $i\in[n]$, be an i.i.d.\ sample of observed outcomes, and write $\eta_0=(\mbox{vec}(A_0),\bm\beta_0^\top)^\top$ for the population vector of coefficients that enters the endpoint linear program \eqref{opt:mnar-with-shadow-matrix-form-theta}. 
We assume that an estimator $\hat\eta_n = (\mbox{vec}(\hat A_n),\hat{\bm\beta}^\top_n)^\top$ of $\eta_0$ based on the observed data is available and satisfies the following weak convergence condition.
\begin{assumption}\label{assmp:convergence-rate-theta}
    We assume the observed data $O_1,\ldots,O_n$ are i.i.d.\ draws from the true data-generating process. There exists a sequence $\tau_n\to\infty$ and a random vector $G\in\mathbb{R}^{|\mathcal{F}|(M+1)}$ such that $\tau_n(\hat\eta_n-\eta_0)$ converges in distribution to a random vector $G$, i.e., $\tau_n(\hat\eta_n-\eta_0) \to_{d} G$.
\end{assumption}

Assumption~\ref{assmp:convergence-rate-theta} is intentionally high-level. It only requires that the first-stage estimates of the probabilities entering the linear program have an asymptotic distribution. For discrete covariates, $\hat\eta_n$ can be formed from empirical frequencies and typically satisfies the assumption with convergence rate $\tau_n=\sqrt n$. For continuous covariates, $\hat\eta_n$ may be obtained from nonparametric or sieve estimators, in which case $\tau_n$ may be slower than $\sqrt n$. The results below apply in either case.


\subsection{A Stable Finite-Sample Estimator}\label{subsec:finite-sample-estimator}

We first describe the estimator for the lower endpoint $\theta_{\min}$. The upper endpoint can be handled analogously by reversing the objective. 
For a generic coefficient vector $\eta=(\operatorname{vec}(A)^\top,\bm\beta^\top)^\top$ and a constant $K>0$, define the localized penalized lower-endpoint value as
\begin{equation}\label{eq:localized-lower-bound}
B_K(\eta)
:=
\bm 1^\top A D\bm 1
+
\min_{0\le \mathbf w\le K\bm 1}
\Bigl\{\bm 1^\top AD\mathbf w + K\|A\mathbf w-\bm\beta\|_1\Bigr\}.
\end{equation}
The first term, $\bm 1^\top A D\bm 1$, is the contribution from observed outcomes. The minimization term represents the contribution from missing outcomes under a candidate completion of the missing data.

The formulation in \eqref{eq:localized-lower-bound} modifies the original LP in \eqref{opt:mnar-with-shadow-matrix-form-theta} in two ways. First, instead of requiring the equality constraint $A\mathbf w=\bm\beta$ to hold exactly, we allow violations of this constraint but penalize them through the term $K\|A\mathbf w-\bm\beta\|_1$. This penalty measures how far a candidate vector $\mathbf w$ is from satisfying the probability restrictions implied by the weak shadow variable.  Second, we restrict the decision vector $\mathbf w$ to the bounded box $[0,K]^M$. This localization rules out unstable solutions with very large weights and makes the value of the penalized LP Lipschitz continuous in the estimated coefficients. Thus, a direct plug-in $B_K(\hat\eta_n)$ would always be a feasible estimator.



We first state a sufficient condition under which the bias for $B_K(\eta)$ is zero, meaning that the localized penalized LP has the same population value as the original LP. The key requirement is that the localization radius $K$ is large enough to contain an optimal primal-dual pair of the population LP. 

\begin{proposition}\label{prop:fix-penalty}
    Suppose there exists an optimal primal-dual solution $(\w^*, \bm\lambda^*)$ of \eqref{opt:mnar-with-shadow-matrix-form-theta} that satisfies $\w^*\in[0, K]^M$ and $\|\bm\lambda^*\|_\infty\le K$. Then the estimator $B_K(\hat\eta_n)$ is consistent for its true estimand, i.e., $B_K(\hat\eta_n)-\theta_{\min}=O_p(\tau_n^{-1})$.
\end{proposition}
To see why this condition is natural, note that the dual representation of \eqref{eq:localized-lower-bound} is
\begin{equation*}
B_K(\eta)
=
\bm 1^\top A D\bm 1
+
\max_{\|\bm\lambda\|_\infty\le K}
\Bigl\{\bm\beta^\top\bm\lambda - K\bigl\|\bigl[A^\top\bm\lambda- DA^\top\bm1 \bigr]_+\bigr\|_1\Bigr\}.
\end{equation*}
The same constant $K$ therefore bounds the size of the dual variables. If there exists an optimal primal-dual solution $(\mathbf w^*,\bm\lambda^*)$ of the original population LP \eqref{opt:mnar-with-shadow-matrix-form-theta} such that $\mathbf w^*\in[0,K]^M$ and $\|\bm\lambda^*\|_\infty\le K$, then the localization box contains a population optimizer and the penalty does not alter the population optimal value. In this case, we would have $B_K(\eta_0)=\theta_{\min}$.

However, setting $K$ to be an unnecessarily large number may not be optimal due to an increase in variance. In particular, an unnecessarily large $K$ can make the estimator more sensitive to sampling noise by enlarging the search region and increasing the penalty level. We therefore seek a radius that is large enough to preserve the population endpoint, but no larger than needed for stable finite-sample performance. The remaining question is how to choose such a radius, which we address next in a data-driven manner. 

\subsubsection{A Data-Driven Algorithm for Selecting $K$.}
This subsection proposes a data-driven procedure that screens a sequence of candidate radii to choose $K$ large enough for the estimator $B_K(\hat\eta_n)$ to remain consistent, yet small enough to keep its variance low. 

The procedure has two inputs. The first is a non-decreasing candidate sequence $ L_1 \le L_2 \le \cdots$ with $L_j\to\infty$, which represents increasingly large localization radii. The second is a diagnostic, called the certification gap, that evaluates whether a candidate radius is large enough to certify the zero-bias condition from Proposition~\ref{prop:fix-penalty}. For a generic coefficient vector $\eta=(\operatorname{vec}(A)^\top,\bm\beta^\top)^\top$, define the certification gap as
\begin{equation}\label{eq:localized-gap}
\Gamma_K(\eta)
:=
\min_{\substack{0\le \mathbf w\le K\bm 1\\ \|\bm\lambda\|_\infty\le K}}
\Bigl\{
 \bm 1^\top AD\mathbf w-\bm\beta^\top\bm\lambda
+2K\|A\mathbf w-\bm\beta\|_1
+2K\bigl\|\bigl[A^\top\bm\lambda- DA^\top\bm1\bigr]_+\bigr\|_1
\Bigr\}.
\end{equation}

The certification gap operationalizes the condition in Proposition~\ref{prop:fix-penalty}. Intuitively, it asks whether there exist primal and dual candidates, both lying within radius $K$, that approximately certify optimality of the original lower-endpoint LP. The two penalty terms measure violations of the primal and dual feasibility conditions. Thus, a small value of $\Gamma_K$ indicates that the radius $K$ is large enough to contain a primal-dual optimal solution. At the population level, the gap is exactly zero if and only if the radius contains an optimal primal-dual pair, which we state in the following Lemma.

\begin{lemma}\label{lem:Gamma-zero-main}
At the true value $\eta_0$, the equality $\Gamma_K(\eta_0)=0$ holds if and only if there exists a primal-dual optimal pair $(\mathbf w^*,\bm\lambda^*)$ for the lower-endpoint LP in \eqref{opt:mnar-with-shadow-matrix-form-theta} such that $\mathbf w^*\in[0,K]^M$ and $\|\bm\lambda^*\|_\infty\le K$. 
\end{lemma}

Lemma~\ref{lem:Gamma-zero-main} provides the basis for selecting $K$ because it shows that $K$ is large enough if and only if $\Gamma_K(\eta_0)=0$. Since the population gap $\Gamma_K(\eta_0)$ is not observable, we evaluate the sample gap $\Gamma_K(\hat\eta_n)$ along the candidate sequence. Let $\delta_n>0$ be a tolerance level for sample size $n$ with $\delta_n\to0$. Then, we select the smallest candidate radius whose sample certification gap is below this tolerance, i.e., we choose 
\[
\widehat{K}_n:=L_{\hat j_n},\;\; \mbox{where}\;\;
\hat j_n:=\min\{1\le j\le n:\Gamma_{L_j}(\hat{\eta}_n)\le \delta_n\}.
\]
If no candidate among $L_1,\ldots,L_n$ satisfies the criterion, we set $\hat j_n=n$ and choose the largest candidate $L_n$. This rule chooses a sufficiently large, but not unnecessarily large, candidate value to remove the localization bias. Our final data-driven estimator for the lower endpoint is then \[\hat{\theta}_{\min} =  B_{\widehat{K}_n}(\hat\eta_n).\]

Next we define the corresponding population radius as $K_*:=\min\{L_j : \Gamma_{L_j}(\eta_0)=0\},$ which is the smallest candidate radius that contains a population optimal primal-dual pair. The next theorem shows that this selection rule consistently recovers the population radius $K_*$ and that the resulting endpoint estimator $B_{\widehat{K}_n}(\hat\eta_n)$ has the same convergence rate as if an appropriate fixed radius had been known in advance.
\begin{theorem}\label{thm:consistency-theta}
Suppose Assumption~\ref{assmp:convergence-rate-theta} holds and the population lower-endpoint LP has at least one finite primal-dual optimal pair. Let the nondecreasing candidate sequence $\{L_j\}_{j=1}^{\infty}$ and thresholds $\{\delta_n\}_{n=1}^{\infty}$ satisfy $L_n\to\infty$, $\delta_n\to0$, $L_n^2/\tau_n\to0$, and $\tau_n\delta_n/L_n^2\to\infty$.
Then $\widehat{K}_n$ converges in probability to $K_*$, and the estimator $\hat{\theta}_{\min} = B_{\widehat K_n}(\hat\eta_n)$ is consistent for $\theta_{\min}$ in the sense that
\[
B_{\widehat K_n}(\hat\eta_n)-\theta_{\min}=O_p(\tau_n^{-1}).
\]
\end{theorem}

Theorem~\ref{thm:consistency-theta} shows that the data-driven choice of $K$ does not affect the first-order convergence rate of the endpoint estimator. The reason is that $\widehat K_n$ converges to the fixed population radius $K_*$, so asymptotically the estimator behaves as if this radius had been known in advance. For example, when the first-stage probabilities are estimated at the usual rate $\tau_n=\sqrt n$, one may take $L_n=n^{1/4-\varepsilon}$ and choose $\delta_n$ so that $L_n^2/n^{1/2}\ll \delta_n\ll1$.

\subsection{Subsampling Inference}\label{subsec:bootstrap}

This subsection develops inference for the endpoints and the full identified set. Because each endpoint is the optimal value of a linear program, it may vary nonsmoothly with the estimated probability constraints. This nonsmoothness can produce non-Gaussian limiting distributions and make the ordinary bootstrap unreliable. We proceed in three steps. First, we characterize the limiting distribution and explain why standard inference can fail. Second, we introduce a subsampling procedure for endpoint inference. Third, we combine one-sided endpoint bounds to construct a confidence region for the full identified set. We focus on $\hat\theta_{\min}=B_{\widehat K_n}(\hat\eta_n)$; the upper-endpoint estimator $\hat\theta_{\max}$ can be handled analogously.

\subsubsection{Asymptotic Distribution of the Endpoint Estimator.} By Theorem~\ref{thm:consistency-theta}, the selected radius $\widehat K_n$ converges in probability to $K_*$. Thus, for first-order inference, the lower-endpoint estimator $\hat\theta_{\min}=B_{\widehat K_n}(\hat\eta_n)$ behaves as if $K_*$ were known. Because $\theta_{\min}=B_{K_*}(\eta_0)$, the asymptotic behavior of the lower-endpoint estimator is governed by the local variation of the value map $B_{K_*}(\cdot)$ around $\eta_0$.

Although localization makes $B_{K_*}(\cdot)$ Lipschitz continuous, the value map need not be fully differentiable. When the population LP has multiple or nearly tied optimal solutions, small perturbations of the estimated probabilities can change the selected optimizer, creating a kink in the value map. The standard delta method therefore does not apply directly. The ordinary bootstrap may also fail because it approximates the local behavior around a sample-selected optimizer rather than the variation induced by switching among population-optimal solutions. We instead use the directional delta method \citep{fang2019inference}, which requires only directional differentiability.
\begin{proposition}\label{prop:limit-distribution}
    Under the conditions of Theorem~\ref{thm:consistency-theta}, we have
    \[ \tau_n(\hat\theta_{\min}-\theta_{\min})=\tau_n\left(B_{\widehat K_n}(\hat\eta_n)-B_{K_*}(\eta_0)\right) \to_{d} B_{K_*}'(G), \]
    where $B_{K_*}'(\cdot)$ is the Hadamard directional derivative of $B_{K_*}(\cdot)$ and $G$ is the limiting random element for $\hat\eta_n$ specified in Assumption \ref{assmp:convergence-rate-theta}. Moreover, $B_{K_*}'(h)$ can be evaluated as the value of a finite-dimensional linear program for every direction $h$. 
\end{proposition}

This proposition characterizes the limiting distribution of the endpoint estimator $\hat\theta_{\min}$. Because $B_{K_*}'(\cdot)$ is generally nonlinear, the limit $B_{K_*}'(G)$ can be non-Gaussian. The proof and the explicit derivative LP are provided in Appendix~\ref{apx:bootstrap-proofs}. 

\subsubsection{Subsampling Procedure.} To approximate this potentially nonstandard distribution, we use recentered subsampling following \citet{politis1994large}. The procedure recomputes the endpoint estimator on smaller samples drawn without replacement and does not require a smooth approximation of the LP value map. Let the subsample size $m$ satisfy $m\to\infty$ with $m/n\to0$. Because subsample fluctuations are larger than the full-sample estimation error, centering each subsample statistic at the full-sample estimate makes the latter error asymptotically negligible. Different subsamples may also select different population-optimal solutions, capturing the optimizer switching that drives the nonsmooth limiting law.


We keep the full-sample radius selector $\widehat K_n$ fixed across subsamples, which is asymptotically valid because $\widehat K_n\to_P K_*$. For each subset $S\subseteq[n]$ with $|S|=m$, let $\hat\eta_{m,S}$ denote the analogue of $\hat\eta_n$ computed from the observations in $S$, and define
\[
T_{m,S}:=\tau_m\Bigl(B_{\widehat K_n}(\hat\eta_{m,S})-B_{\widehat K_n}(\hat\eta_n)\Bigr).
\]
The statistic $T_{m,S}$ is the subsampling analogue of the centered full-sample statistic $\tau_n(\hat\theta_{\min}-\theta_{\min})$. 
The corresponding subsampling distribution and its $\gamma$-quantile are
\[
\hat J_{n,m}(t)
:=
\binom{n}{m}^{-1}\sum_{|S|=m}\bm 1\{T_{m,S}\le t\},
\qquad
\hat q_{n,m}(\gamma)
:=
\inf\{t\in\R:\hat J_{n,m}(t)\ge \gamma\}.
\]
In practice, $\hat J_{n,m}$ can be approximated by drawing a large number of subsamples. This approximation does not affect the asymptotic result provided that the number of draws diverges. We use the empirical quantiles to construct the following two-sided confidence interval for $\theta_{\min}$:
\begin{equation}\label{eq:fs-endpoint-ci}
\widehat{\mathrm{CI}}_{1-\alpha}
:=
\left[
\hat\theta_{\min, n}-\tau_n^{-1}{\hat q_{n,m}(1-\alpha/2)},
\ \hat\theta_{\min, n}-\tau_n^{-1}{\hat q_{n,m}(\alpha/2)}
\right].
\end{equation}
The interval in \eqref{eq:fs-endpoint-ci} recenters the subsampling quantiles around the full-sample estimator $\hat\theta_{\min, n}$. Intuitively, it uses the estimated distribution of the $\tau_m$-scaled error to determine how far the interval must extend on either side of $\hat\theta_{\min, n}$ to achieve asymptotically valid coverage. 

The following theorem formalizes this procedure by applying the large-sample subsampling result of \citet[Theorem~3.1]{politis1994large} to our selected-radius endpoint statistic.
\begin{theorem}\label{thm:subsample-bootstrap}
Under the conditions of Proposition~\ref{prop:limit-distribution}, suppose $m\to\infty$, $m/n\to0$, and $\tau_m/\tau_n\to0$. Let $\mathcal{L}$ denote the limit distribution of $\tau_n(\hat\theta_{\min,n}-\theta_{\min})$ as in Proposition \ref{prop:limit-distribution}, and let $J_{\mathcal{L}}$ denote its cdf. Then:
\begin{enumerate}
    \item[(a)] For every continuity point $t$ of $J_\mathcal{L}$, we have $ \hat J_{n,m}(t)\to_P J_\mathcal{L}(t)$.
    \item[(b)] Let $q_\mathcal{L}(\gamma):=\inf\{t\in\R:J_\mathcal{L}(t)\ge \gamma\}$. If, for every $\varepsilon>0$,
    \[
    J_\mathcal{L}\bigl(q_\mathcal{L}(\gamma)-\varepsilon\bigr)<\gamma<J_\mathcal{L}\bigl(q_\mathcal{L}(\gamma)+\varepsilon\bigr),
    \]
    then $\hat q_{n,m}(\gamma)\to_P q_\mathcal{L}(\gamma)$.
    \item[(c)] If the condition in part (b) holds for $\gamma=\alpha/2$ and $\gamma=1-\alpha/2$, and $J_\mathcal{L}$ is continuous at both corresponding quantiles, then $\P\bigl(\theta_{\min}\in \widehat{\mathrm{CI}}_{1-\alpha}\bigr)\to 1-\alpha$.
\end{enumerate}
\end{theorem}

Theorem~\ref{thm:subsample-bootstrap} shows that subsampling consistently estimates the nonstandard limit law of the lower-endpoint estimator. Part (a) establishes consistency of the subsampling cdf, part (b) converts this into consistency of the critical values, and part (c) gives asymptotically valid coverage for $\theta_{\min}$. The proof first uses $\widehat K_n\to_P K_*$ to reduce the analysis of the statistic to one based on a fixed-radius value map and then verifies the conditions of \citet[Theorem~3.1]{politis1994large}. Importantly, this inference procedure does not require a stable basis, a smooth value map, or any rank condition on the matrix $A_0$, which contrasts with what is often needed in LP-based inference \citep{goff2024inference, voronin2025linear}. Common choices such as $m=\lfloor n^{2/3}\rfloor$ or $m=\lfloor n^{3/4}\rfloor$ satisfy the required rate conditions when $\tau_n=\sqrt n$. Although the subsampling quantiles are estimated at the subsample scale, their contribution to the confidence interval is multiplied by $\tau_n^{-1}$ in \eqref{eq:fs-endpoint-ci}, so this additional approximation error is asymptotically negligible.

\subsubsection{Confidence Region for the Identified Set.} Applying subsampling to both endpoints yields a confidence region for the identified set $\Theta_0=[\theta_{\min},\theta_{\max}]$. Let $\hat q_{n,m}^{L}$ and $\hat q_{n,m}^{U}$ denote the subsampling quantiles for the lower and upper endpoints, respectively. Combining a one-sided lower confidence bound for $\theta_{\min}$ with a one-sided upper confidence bound for $\theta_{\max}$ gives
\begin{equation}\label{eq:fs-region-ci}
\widehat{\mathrm{CR}}_{1-\alpha}
:=
\left[
\hat\theta_{\min,n}-\tau_n^{-1}\hat q_{n,m}^{L}(1-\alpha/2),\,
\hat\theta_{\max,n}-\tau_n^{-1}\hat q_{n,m}^{U}(\alpha/2)
\right].
\end{equation}
\begin{proposition}[Confidence Region for the Identification Set]\label{prop:ci-identification-region}
Under the conditions of Theorem~\ref{thm:subsample-bootstrap} for both endpoint estimators, suppose the lower and upper endpoint limit laws are continuous at the quantiles used in \eqref{eq:fs-region-ci}. Then
\[
\liminf_{n\to\infty}\P\bigl(\Theta_0\subseteq \widehat{\mathrm{CR}}_{1-\alpha}\bigr)\ge 1-\alpha.
\]
\end{proposition}
Proposition~\ref{prop:ci-identification-region} follows from the one-sided endpoint coverage statements and Bonferroni's inequality. The construction is conservative only through this final Bonferroni step, which protects simultaneous coverage of the two endpoints.

\begin{remark}[Other Population Quantities]\label{rem:general-objective}
    Although this section focuses on inference for the mean outcome, the same finite-sample estimator and subsampling procedure apply more generally. Specifically, the term $\bm 1^\top A D\mathbf w$ can be replaced by a generic objective $c^\top\mathbf w$, where the vector $c$ encodes the population quantity of interest, e.g., $\mathbb E[g(Y)]$. Then the localized penalized LP and the subsampling inference procedure apply without modification.
\end{remark}

\subsection{Continuous Covariates}\label{subsec:continuous-covariates}
The preceding analysis formulates the finite-sample procedure within a fixed covariate stratum. This notation is convenient when $X$ has finite support, because the constraints can be stacked across strata. When $X$ is continuously distributed, the same identification logic still applies pointwise in $x$, but solving a separate LP at every covariate value serves only as a conceptual representation. A more useful formulation is to treat the unknown response odds as a measurable function of $X$ and to express the pointwise restrictions first as conditional and then as unconditional moment equalities. In fact, the LP in \eqref{opt:mnar-with-shadow-matrix-form-theta} can be represented as a moment-equality model, as shown in the following proposition.

\begin{proposition}\label{prop:continuous-x-cmom}
Suppose regular conditional probabilities exist and $\pi_x(y)\in(0,1]$ for $P_{Y,X}$-almost every $(y,x)$. For each $f\in\mathcal{F}$, define the observable moment function
\[
m_f(O;w)
:=\mathbf{1}\{R=0,F=f\}
-\sum_{y=1}^M \mathbf{1}\{R=1,F=f,Y=y\}\,w_X(y),
\]
where $O=(R,RY,F,X)$ is the observed variable. Then the constraint in \eqref{opt:mnar-with-shadow-matrix-form-theta} can be written as
\begin{equation}\label{eq:cont-x-cmom}
\E\big[m_f(O;w)\mid X\big]=0
\qquad\text{a.s.\ for all }f\in\mathcal{F}.
\end{equation}
\end{proposition}

Proposition~\ref{prop:continuous-x-cmom} shows that the linear constraint can be written as conditional moment equalities. When $X$ is continuous, however, this representation yields infinitely many linear constraints, complicating finite-sample inference. One can construct a finite number of constraints by noting that, for any function $q(x)$, constraint \eqref{eq:cont-x-cmom} implies that $\E\big[m_f(O;w)q(X)\big]=0$ for all $f\in\mathcal{F}$. Thus, one can choose a finite class of functions $\mathcal Q$ and calculate the identification region under the following system of unconditional moment equalities:
\begin{equation}\label{eq:cont-x-ucmom}
    \E\big[m_f(O;w)q(X)\big]=0
\qquad \forall f\in\mathcal{F}, q\in\mathcal{Q}.
\end{equation}
If $\mathcal Q$ is rich enough to characterize conditional mean zero, then the identification region from \eqref{eq:cont-x-ucmom} coincides with that from \eqref{eq:cont-x-cmom}. Choosing $q(X)$ to be cell indicators recovers the discrete stratification approach; choosing basis or sieve functions yields increasingly refined approximations without requiring a separate LP for every covariate value.

\section{Experiments}\label{sec:experiment}
We evaluate the proposed identification bounds and the localized penalized estimator on both simulated data and real-world customer-service dialogue data. In Section \ref{subsec:simulations}, we use simulations to verify the theoretical results in Section~\ref{sec:set-expansion}, including the $\tau_n$-rate of convergence of the proposed estimator, the validity of subsampling, and the failure of the ordinary bootstrap in this setting. In Section \ref{subsec:semi-syn}, we conduct semi-synthetic experiments on real customer-service dialogue data to study LLM-based weak shadow variables and compare the resulting bounds with standard MNAR methods. We show that, with suitable shadow-variable design, our method achieves lower estimation error than the point-estimation baselines and tighter identification regions than the no-shadow-variable benchmark.

\subsection{Simulations}\label{subsec:simulations}
\begin{figure}[htbp]
    \centering
    \includegraphics[width=0.86\textwidth]{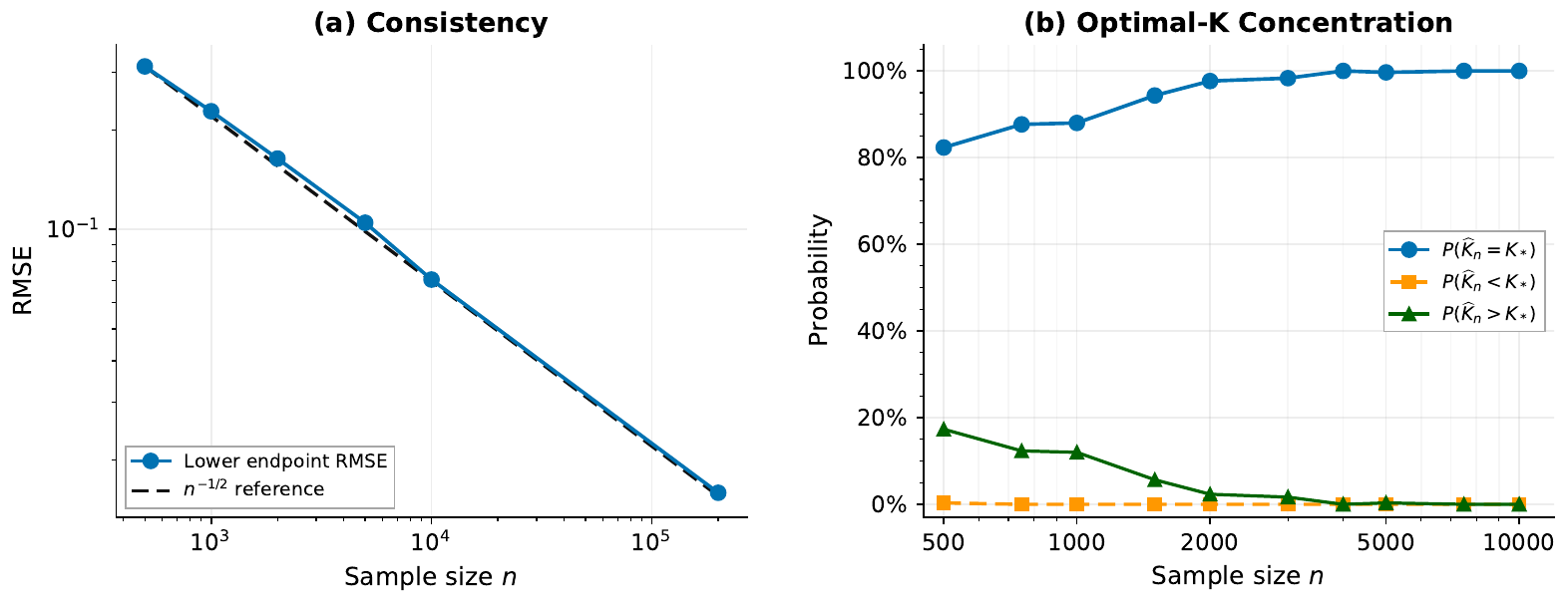}
    \caption{Finite-sample estimation of the lower endpoint. Panel (a) plots RMSE against $n$ on log-log scales with an $n^{-1/2}$ reference line. Panel (b) reports selection probabilities for the localized tuning index $K$, with population target $K_*=10$.}
    \label{fig:localized-gap-estimation}
\end{figure}

We first use simulated data to verify the estimation and inference theory in Section~\ref{sec:set-expansion}. We simulate a platform rating setting with $M=6$, so customers rate on a scale from 1 to 6. Outcomes are MNAR, with response probabilities that vary across outcome values $Y$, and we construct a binary shadow variable that depends only on the actual rating. The data-generating process is reported in Table~\ref{tab:localized-gap-dgp}.
\begin{table}[H]
  \centering
  \begin{tabular}{lcccccc}
  \toprule
   & $Y=1$ & $Y=2$ & $Y=3$ & $Y=4$ & $Y=5$ & $Y=6$ \\
  \midrule
  $P(Y)$ & 0.05 & 0.10 & 0.20 & 0.30 & 0.25 & 0.10 \\
  $P(F=1 \mid Y)$ & 0.05 & 0.20 & 0.35 & 0.50 & 0.65 & 0.70 \\
  $P(R=1 \mid Y)$ & 0.90 & 0.60 & 0.25 & 0.20 & 0.55 & 0.90 \\
  \bottomrule
\end{tabular}
\caption{Simulation setup.}\label{tab:localized-gap-dgp}
\end{table}
Under this setting, most customers have ratings from 3 to 5 and only a few have extreme ratings. Response propensity is U-shaped, so customers with more extreme ratings are more likely to respond. The shadow variable is positively correlated with the actual rating because higher ratings are more likely to generate $F$ with value $1$. In this design, the lower and upper endpoints of the identification region are $\theta_L=3.893$ and $\theta_U=4.132$, while the true full-data mean is $3.900$. For estimation, we choose $L_n=5\lfloor n^{1/5}\rfloor$ and $\delta_n=0.01/\log n$, yielding the population target $K_*=10$. We report the result for the lower endpoint, and the upper endpoint is analogous. The identifiable coefficients are simple sample averages that are root-$n$ consistent. Figure~\ref{fig:localized-gap-estimation} plots estimator consistency and selection of $K$.

Panel (a) of Figure~\ref{fig:localized-gap-estimation} shows that the localized estimator converges at the parametric rate, with the log-log RMSE curve essentially matching the $n^{-1/2}$ benchmark. Panel (b) shows that the data-driven selector concentrates rapidly on the target localization level with $\P(\widehat{K}_n=K_*)$ rising from $0.823$ at $n=500$ to $1.000$ at $n=10{,}000$, which matches the selector-stability theory in Section \ref{sec:set-expansion}.

\begin{figure}[htbp]
    \centering
    \includegraphics[width=0.86\textwidth]{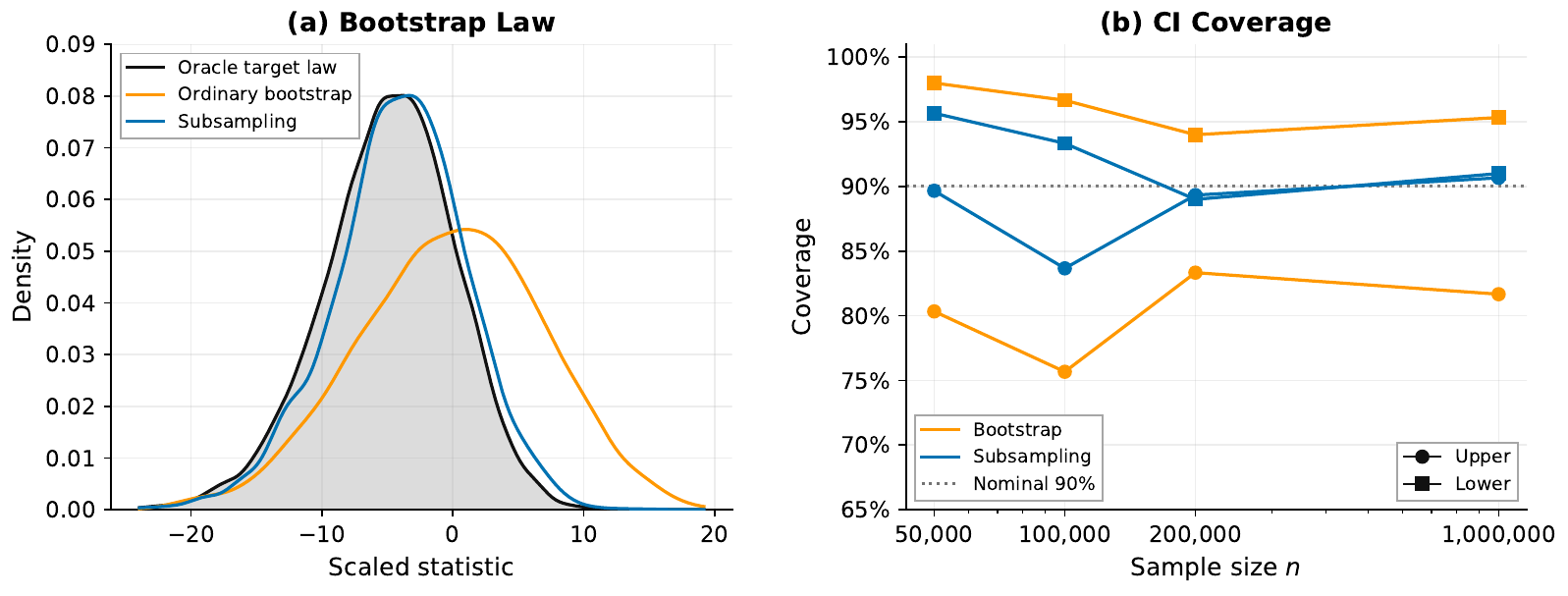}
    \caption{Lower-endpoint inference. Panel (a) compares the oracle root-$n$ law with ordinary bootstrap and subsampling approximations. Panel (b) reports empirical coverage of one-sided $90\%$ confidence intervals; colors denote the resampling method and markers distinguish upper and lower confidence bounds.}
    \label{fig:localized-gap-inference}
\end{figure}

Next, we examine inference. Inference is challenging here because the set of optimal solutions to the identification LP is a three-dimensional face rather than a single point. Panel (a) of Figure~\ref{fig:localized-gap-inference} compares the lower-endpoint oracle law with ordinary bootstrap and subsampling approximations using $n=1{,}000{,}000$, subsample size $m=5{,}000$, and $5{,}000$ resampling draws. Panel (b) reports coverage at $n\in\{50{,}000,100{,}000,200{,}000,1{,}000{,}000\}$ using the corresponding subsample sizes $m\in\{1{,}500,2{,}500,5{,}000,5{,}000\}$, 300 Monte Carlo replications, and 399 resampling draws per replication. We can see from Panel (a) that the oracle limit distribution is non-normal, and the ordinary bootstrap is badly mis-centered because the endpoint is a nonsmooth value function with nonunique population optimizers. By contrast, subsampling corrects this bias and closely tracks the oracle law. Panel (b) translates these differences into coverage in a sense that the ordinary-bootstrap upper confidence bound under-covers at 82\%, whereas its lower confidence bound over-covers at 95\%. Subsampling approaches the 90\% target for both one-sided bounds.

\subsection{Semi-Synthetic Experiments}\label{subsec:semi-syn}
Next, we validate our estimator using a semi-synthetic experiment based on real dialogue satisfaction data from e-commerce customer-service interactions. We simulate a realistic MNAR structure and a range of candidate shadow variables to evaluate estimation performance.

\subsubsection{Data.}
We use the public User Satisfaction Simulation (USS) dataset \citep{sun2021simulating}, a collection of human-annotated dialogues compiled from five public dialogue corpora. We focus on the JD.com corpus, which comes from a major online retailer in China and contains 3,300 Chinese customer-service dialogues. Each dialogue contains multiple conversation turns between users and automated systems; two examples are provided in Appendix~\ref{app:data-examples}. Independent human annotators provide overall satisfaction ratings on a 1--5 scale, which we will use as ground truth. Table~\ref{tab:jddc-distribution} reports the distribution in the JDDC subset, which is concentrated at ratings 3 and 4.

\begin{table}[htbp]
\centering
\caption{Distribution of annotator ratings in the JDDC subset of the USS dataset ($n=3{,}300$)}
\label{tab:jddc-distribution}
\begin{tabular}{lccccc}
\toprule
Rating & 1 & 2 & 3 & 4 & 5 \\
\midrule
Count & 2 & 144 & 725 & 2287 & 142 \\
Percentage & 0.1\% & 4.4\% & 22.0\% & 69.3\% & 4.3\% \\
\bottomrule
\end{tabular}
\end{table}

Since all ratings are observed in the original data, we simulate response behavior using LLM digital twins. For each dialogue, we randomly draw a persona from Twin-2k-500 \citep{toubia2025database}, which consists of 2,000 persona summaries, and ask an LLM acting as that person to report a probability of leaving a rating and a single simulated response indicator. In the downstream estimator comparison, we generate repeated response indicators as Bernoulli draws with the reported probability as their mean. This design allows response propensity to depend on both the realized outcome and the simulated persona, mimicking voluntary customer feedback. We use \texttt{GPT-5.4-mini}, \texttt{Grok-4-20-Reasoning}, and \texttt{DeepSeek-V3.2} to simulate responses for robustness.

\subsubsection{Constructing Weak Shadow Variables from LLMs.} We construct weak shadow variables by prompting LLMs to evaluate the dialogue from different perspectives. The design goal is to obtain variables that satisfy the exclusion assumption, Assumption~\ref{ass:cond-indep}, so that the constructed variable contains no additional information about the response decision after conditioning on the actual customer rating.

Our prompt design is a $3 \times 3$ matrix that crosses two dimensions: \emph{perspective} and \emph{construct}. The perspective axis varies who evaluates the interaction, including the exact customer represented by a digital twin, a generic customer facing the same problem, or a neutral service auditor. The construct axis varies what is measured, which includes overall rating, goal completion, or handling quality. We also include a binary branch with narrower yes/no goal-completion questions, allowing us to compare ordinal and binary outputs within the goal-completion construct. Only digital-twin prompts receive persona summaries; all other prompts use dialogue alone. Table~\ref{tab:prompt-design-matrix} summarizes the prompt families, each run through \texttt{GPT-5.4-mini}, \texttt{GPT-4.1-mini}, and \texttt{Grok-4-20-Reasoning}.

\begin{table}[htbp]
\centering
\small
\caption{Design matrix for the systematic prompt experiment. The ordinal columns return $1$--$5$ scores; the binary branch returns a yes/no judgment coded as $0/1$. Only the digital-twin row uses persona summaries; all other prompts use dialogue alone.}
\label{tab:prompt-design-matrix}
\setlength{\tabcolsep}{3.3pt}
\makebox[\textwidth][c]{%
\begin{tabular}{>{\raggedright\arraybackslash}p{1.90cm}
                >{\raggedright\arraybackslash}p{3.00cm}
                >{\raggedright\arraybackslash}p{3.20cm}
                >{\raggedright\arraybackslash}p{3.25cm}
                >{\raggedright\arraybackslash}p{3.55cm}}
\toprule
Perspective & \multicolumn{3}{c}{Ordinal main matrix ($1$--$5$)} & Binary branch ($0/1$) \\
\cmidrule(lr){2-4}\cmidrule(l){5-5}
 & Overall rating & Goal completion & Handling quality & Goal completion \\
\midrule
Digital twin user & How would \emph{you} rate this interaction? & How much did this solve your problem or give a workable next step? & How well did the system handle your request? & \textbf{Yes/No:} Is the issue resolved, or at least far enough along to proceed? \\
Generic user & How would a typical customer rate this interaction? & How successful was this for a typical customer with the same problem? & How well would a typical customer judge the handling? & \textbf{Yes/No:} Would a typical customer consider the issue resolved enough to proceed? \\
Service auditor & How strong was the interaction overall from a service perspective? & How much did the interaction achieve resolution or useful closure? & How well did the system understand and manage the request? & \textbf{Yes/No:} Was the issue resolved or clearly moved to a workable next step? \\
\bottomrule
\end{tabular}
\par}
\end{table}

\subsubsection{Structural Results for Prompt Design.} We evaluate each prompt along two dimensions. Exclusion corresponds to Assumption~\ref{ass:cond-indep} and is required for a valid weak shadow variable, whereas relevance captures the association between $F$ and $Y$. A stronger association can help tighten the identification region, and we use the Spearman correlation between $F$ and $Y$ as an empirical measure of relevance. In this semi-synthetic diagnostic, we take $X$ to be empty, so the exclusion restriction reduces to $F\indep R\mid Y$. We assess exclusion using two complementary diagnostics: a stratified likelihood-ratio test of $F\indep R\mid Y$ based on the realized response indicator and the within-$Y$ spread in the simulated response propensity across values of $F$. The first diagnostic measures statistical evidence against the exclusion restriction, while the second measures the magnitude of the departure using the response probabilities available in our semi-synthetic design. Appendix~\ref{app:prompt-diagnostics} provides the formal definitions and aggregation procedure.

\begin{figure}[htbp]
    \centering
    \includegraphics[width=0.86\textwidth]{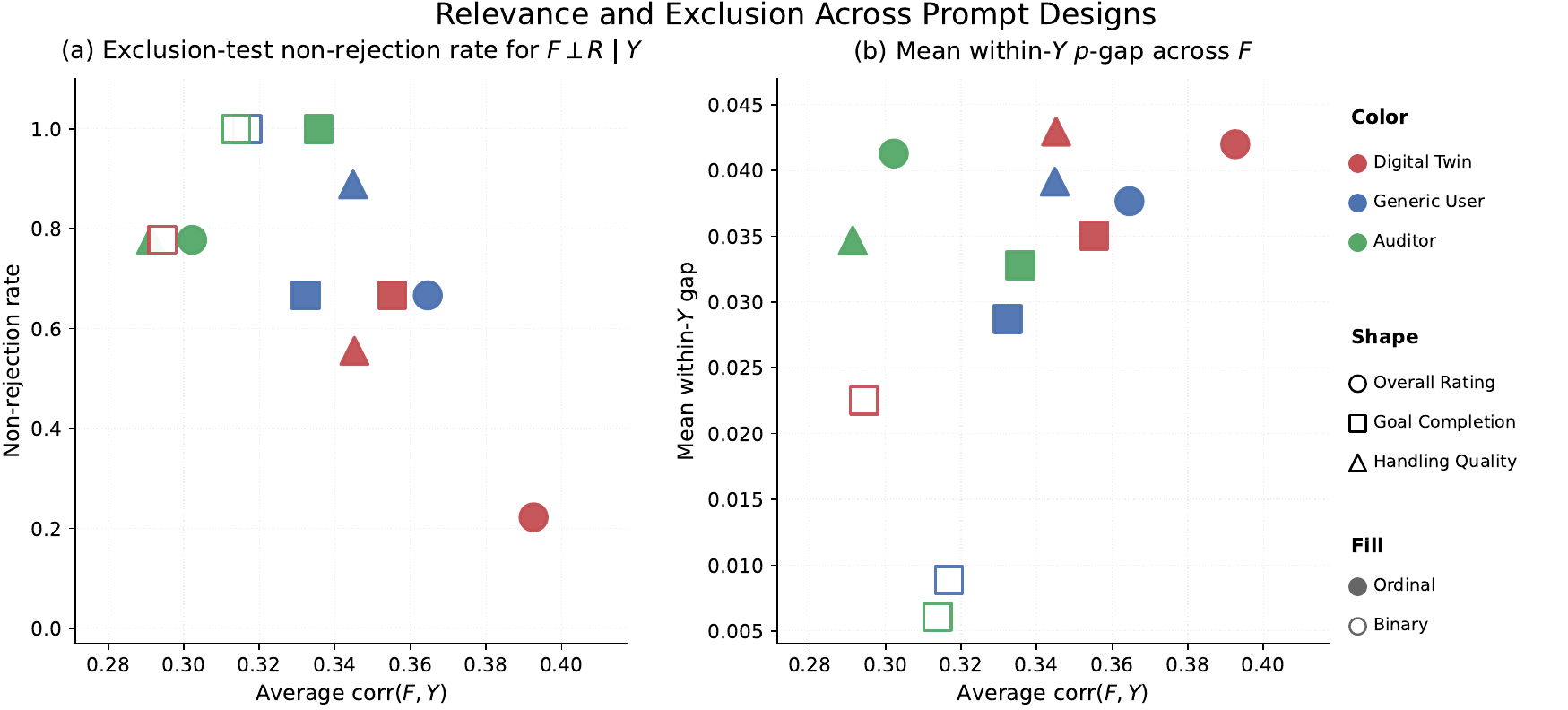}
    \caption{Prompt-level relevance-exclusion frontier. The left panel plots $\mathrm{Spearman}(F,Y)$ against the non-rejection rate of the exclusion test for $F \indep R \mid Y$. The right panel plots the same relevance measure against the mean within-$Y$ spread in $\mathbb{E}[R \mid F,Y]$. Each point averages over valid $F$-model and $R$-model pairings.}
    \label{fig:prompt-relevance-independence}
\end{figure}

Figure~\ref{fig:prompt-relevance-independence} demonstrates the results. First, all shadow variables are positively associated with the outcome, with average Spearman correlations ranging from 0.29 to 0.39. Second, all prompt-level within-$Y$ response-propensity gaps are below 0.045, suggesting that departures from the exclusion restriction are modest in magnitude and providing an empirical scale for calibrating the sensitivity analysis. Finally, the figure reveals a trade-off between relevance and exclusion. Prompts that contain persona information may be more relevant for predicting the rating, but they are also more likely to capture user characteristics that directly affect response propensity. These findings are also robust to the choice of language model, as shown in Appendix \ref{app:model-robustness}.


Overall, the evidence suggests that the most credible weak shadow variables come not from prompts that most closely mimic the customer's own rating, but from prompts that are depersonalized, task-focused, and relatively factual. In particular, auditor-style goal-completion and binary-resolution prompts emerge as the strongest candidates because they remain informative about $Y$ while more plausibly satisfying the exclusion restriction. This echoes our view of LLMs as external evaluators of user experience rather than simulators of user behavior.

\subsubsection{Downstream Estimation Results}
\begin{figure}[htbp]
    \centering
    \includegraphics[width=0.86\textwidth]{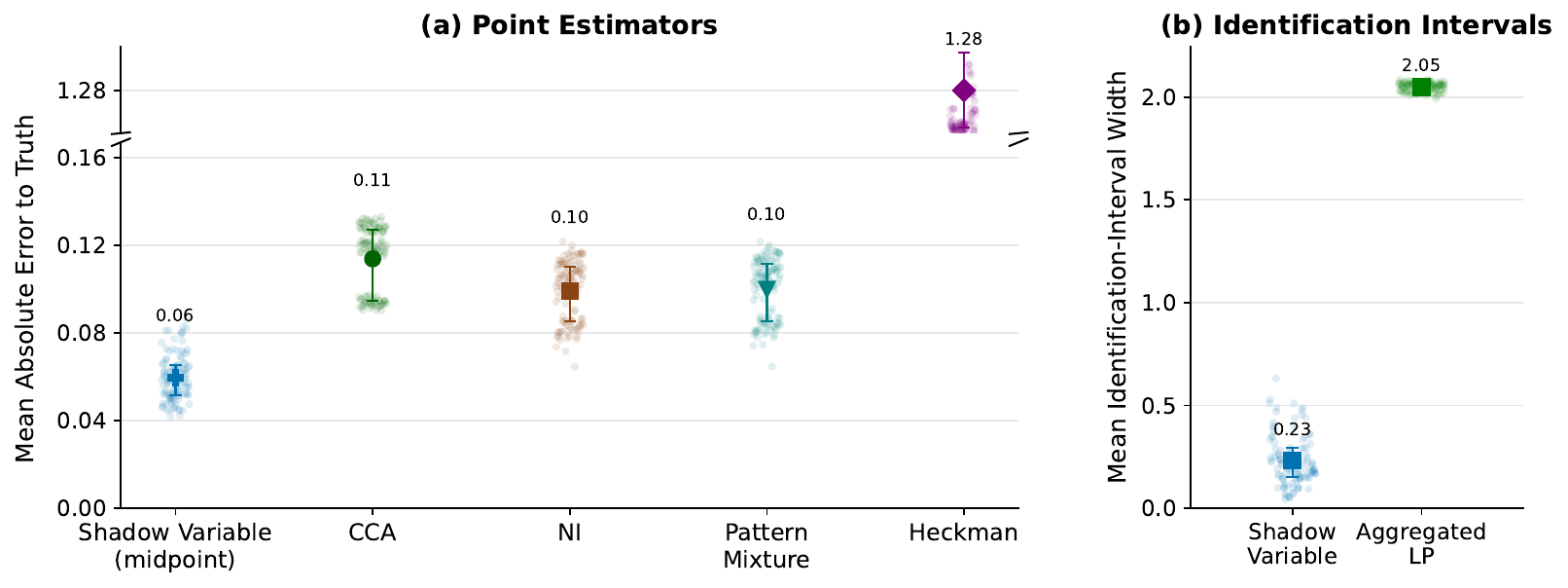}
    \caption{Estimator comparison in the semi-synthetic MNAR experiment. Each faint dot is one structural-cell average over $100$ Bernoulli resamples, the large marker is the mean across cells, and whiskers show the interquartile range. Panel (a) compares the midpoint of the proposed interval with point estimators using absolute error; panel (b) compares the widths of the proposed and aggregated-LP identification intervals.}
    \label{fig:method-comparison-main}
\end{figure}

Finally, we test how much the extracted shadow variables improve downstream estimation of the mean outcome. We compare our method with the following baselines:
\begin{itemize}
\item \textbf{Complete Case Analysis (CCA)}: averages only the observed outcomes.
\item \textbf{Naive Imputation (NI)}: fits a linear regression of $Y$ on $F$ among respondents, imputes missing outcomes using the fitted conditional mean, and averages over all units.
\item \textbf{Heckman Selection Model (Heckman)} \citep{heckman1979sample}: models selection via probit regression on $F$ and corrects for selection bias using the inverse Mills ratio, assuming joint normality of outcome and selection errors.
\item \textbf{Pattern-Mixture Model (PM)} \citep{rubin1987calculation, little1994class}: imputes missing outcomes using respondent conditional means given $F$, imposing $\E[Y\mid F,R=0]=\E[Y\mid F,R=1]$.
\item \textbf{Aggregated LP}: computes the identification region without shadow variables, as in Section~\ref{subsec:partial-identification-simplified}.
\end{itemize}
Here, the NI, Heckman, and PM methods use $F$ as an additional covariate in the model. Binary prompt outputs are remapped from $\{0,1\}$ to $\{1,5\}$ before estimation. For each prompt family, prompt-side LLM, and missingness simulator, we run $100$ Bernoulli resamples of the response indicator, yielding $12 \times 3 \times 3 = 108$ structural cells. Panel (a) of Figure~\ref{fig:method-comparison-main} reports absolute error for the midpoint of our interval and for the point estimators. Panel (b) compares the mean width of our interval with that of the aggregated-LP interval.

As shown in Figure~\ref{fig:method-comparison-main}, the midpoint of the proposed shadow-variable interval achieves mean absolute error $0.06$, substantially below all point-estimation baselines shown in panel (a). At the set level, the shadow-variable estimator has a mean interval width of $0.23$, compared with $2.05$ for the aggregated LP benchmark, representing an average reduction of approximately $89\%$. These results illustrate the value of using a shadow variable to shrink the identification region, beyond merely treating the LLM output as another covariate in NI or pattern-mixture models.

\begin{figure}[htbp]
    \centering
    \includegraphics[width=0.86\textwidth]{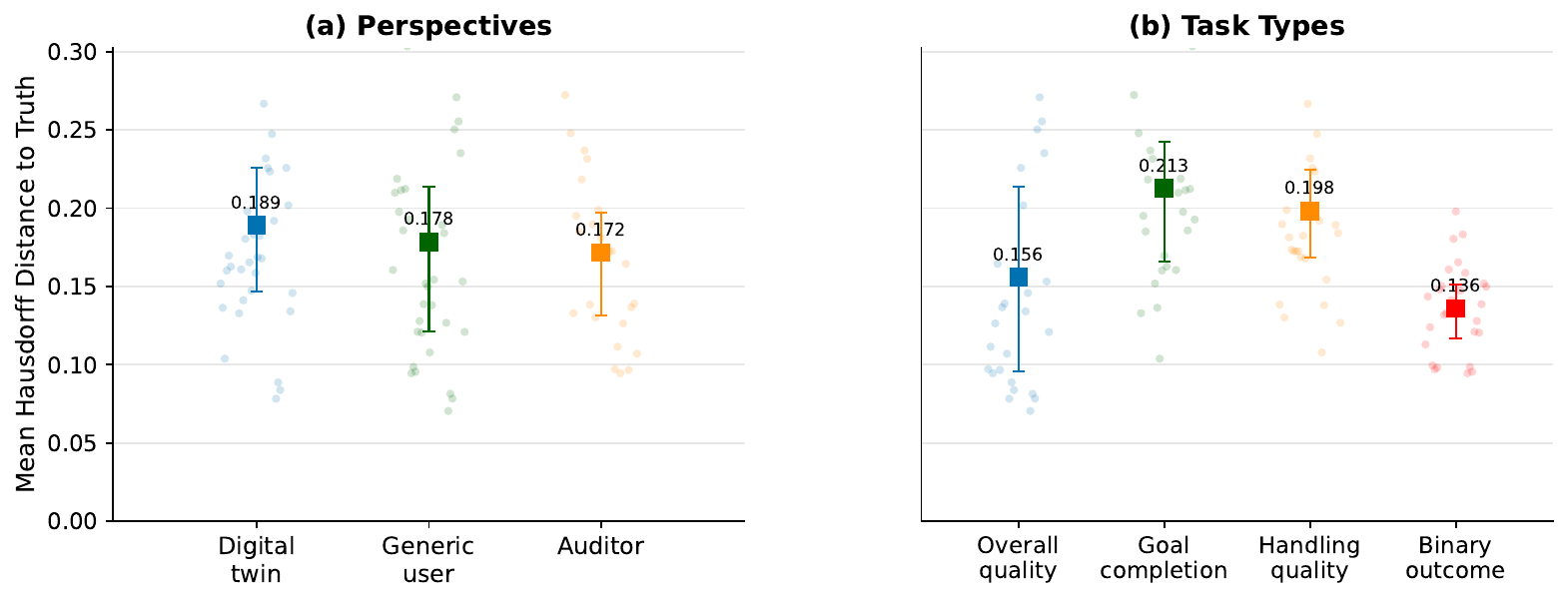}
    \caption{Prompt-design decomposition of identification-interval width. Each faint dot is one structural-cell average for the shadow-variable estimator, the large marker is the mean across cells, and whiskers show the interquartile range. Panel (a) compares prompt perspectives; panel (b) compares task framings and output types, including the binary branch.}
    \label{fig:prompt-comparison-main}
\end{figure}

Figure~\ref{fig:prompt-comparison-main} shows that this improvement is systematic under the prompt groupings used in the figure. Auditor prompts achieve the smallest mean interval width ($0.222$), followed by generic-user prompts ($0.240$) and digital-twin prompts ($0.255$). Across the task and output-type groups, binary-output prompts perform best at $0.162$, followed by overall-quality prompts at $0.196$, whereas handling-quality and ordinal goal-completion prompts are weaker at $0.264$ and $0.307$. These downstream results reinforce the structural finding that useful weak shadow variables tend to use a depersonalized perspective and a narrower, more factual task framing.

\section{Conclusion}\label{sec:conclusion}

We study the problem of estimating population quantities when outcomes are missing not at random, a pervasive challenge on digital platforms and in social surveys. Rather than imposing strong parametric assumptions or seeking point identification under potentially fragile conditions, we adopt a partial identification perspective and show that sharp bounds on the mean can be computed via a pair of linear programs. Our key insight is that measurements from pretrained models---including large language models---can be incorporated as weak shadow variables to tighten these bounds. Unlike classical shadow-variable approaches, our framework does not require completeness or strong measurement accuracy; it extracts useful information from imperfect measurements while providing valid coverage guarantees through a local penalized estimator. Our simulations demonstrate valid coverage across diverse missingness patterns. In the semi-synthetic customer-service experiment, even simple LLM measurements reduce the average width of the identification intervals by approximately 89\% relative to the no-shadow-variable benchmark.


{
\bibliographystyle{informs2014}
\bibliography{ref}

\begin{thebibliography}{54}
\providecommand{\natexlab}[1]{#1}
\providecommand{\url}[1]{\texttt{#1}}
\providecommand{\urlprefix}{URL }

\bibitem[{Abrevaya \protect\BIBand{} Donald(2017)}]{abrevaya2017gmm}
Abrevaya J, Donald SG (2017) A gmm approach for dealing with missing data on
  regressors. \emph{Review of Economics and Statistics} 99(4):657--662.

\bibitem[{Angelopoulos et~al.(2023{\natexlab{a}})Angelopoulos, Bates,
  Fannjiang, Jordan, \protect\BIBand{} Zrnic}]{angelopoulos2023prediction}
Angelopoulos AN, Bates S, Fannjiang C, Jordan MI, Zrnic T (2023{\natexlab{a}})
  Prediction-powered inference. \emph{Science} 382(6671):669--674.

\bibitem[{Angelopoulos et~al.(2023{\natexlab{b}})Angelopoulos, Duchi,
  \protect\BIBand{} Zrnic}]{angelopoulos2023ppi++}
Angelopoulos AN, Duchi JC, Zrnic T (2023{\natexlab{b}}) Ppi++: Efficient
  prediction-powered inference. \emph{arXiv preprint arXiv:2311.01453} .

\bibitem[{Beresteanu \protect\BIBand{}
  Molinari(2008)}]{beresteanu2008asymptotic}
Beresteanu A, Molinari F (2008) Asymptotic properties for a class of partially
  identified models. \emph{Econometrica} 76(4):763--814.

\bibitem[{Bollinger et~al.(2019)Bollinger, Hirsch, Hokayem, \protect\BIBand{}
  Ziliak}]{bollinger2019trouble}
Bollinger CR, Hirsch BT, Hokayem CM, Ziliak JP (2019) Trouble in the tails?
  what we know about earnings nonresponse 30 years after lillard, smith, and
  welch. \emph{Journal of Political Economy} 127(5):2143--2185.

\bibitem[{Brand et~al.(2024)Brand, Israeli, \protect\BIBand{}
  Ngwe}]{brand2024using}
Brand J, Israeli A, Ngwe D (2024) Using gpt for market research.
  \emph{Proceedings of the 25th ACM Conference on Economics and Computation},
  613--613.

\bibitem[{Chen et~al.(2025)Chen, Ao, \protect\BIBand{}
  Simchi-Levi}]{chen2025utilizing}
Chen H, Ao R, Simchi-Levi D (2025) Utilizing external predictions for data
  collection: Joint optimization of sampling and measurement. \emph{Available
  at SSRN 5025010} .

\bibitem[{Chernozhukov et~al.(2007)Chernozhukov, Hong, \protect\BIBand{}
  Tamer}]{ChernozhukovHongTamer2007}
Chernozhukov V, Hong H, Tamer E (2007) Estimation and confidence regions for
  parameter sets in econometric models. \emph{Econometrica} 75(5):1243--1284,
  \urlprefix\url{http://dx.doi.org/10.1111/j.1468-0262.2007.00794.x}.

\bibitem[{Das et~al.(2003)Das, Newey, \protect\BIBand{}
  Vella}]{das2003nonparametric}
Das M, Newey WK, Vella F (2003) Nonparametric estimation of sample selection
  models. \emph{The Review of Economic Studies} 70(1):33--58.

\bibitem[{Dell \protect\BIBand{} Rambachan(2026)}]{dell2026measurement}
Dell M, Rambachan A (2026) The measurement revolution? credible measurement and
  inference in the age of ai .

\bibitem[{Dellarocas \protect\BIBand{} Wood(2008)}]{dellarocas2008sound}
Dellarocas C, Wood CA (2008) The sound of silence in online feedback:
  Estimating trading risks in the presence of reporting bias. \emph{Management
  science} 54(3):460--476.

\bibitem[{d’Haultfoeuille(2010)}]{d2010new}
d’Haultfoeuille X (2010) A new instrumental method for dealing with
  endogenous selection. \emph{Journal of Econometrics} 154(1):1--15.

\bibitem[{Fang \protect\BIBand{} Santos(2019)}]{fang2019inference}
Fang Z, Santos A (2019) Inference on directionally differentiable functions.
  \emph{The Review of Economic Studies} 86(1):377--412.

\bibitem[{Fay(1986)}]{fay1986causal}
Fay RE (1986) Causal models for patterns of nonresponse. \emph{Journal of the
  American Statistical Association} 81(394):354--365.

\bibitem[{Gao et~al.(2025)Gao, Lee, Burtch, \protect\BIBand{}
  Fazelpour}]{gao2025take}
Gao Y, Lee D, Burtch G, Fazelpour S (2025) Take caution in using llms as human
  surrogates. \emph{Proceedings of the National Academy of Sciences}
  122(24):e2501660122.

\bibitem[{Goff \protect\BIBand{} Mbakop(2025)}]{goff2024inference}
Goff L, Mbakop E (2025) Inference on the value of a linear program.
  \emph{Working paper} .

\bibitem[{Goli \protect\BIBand{} Singh(2024)}]{goli2024frontiers}
Goli A, Singh A (2024) Frontiers: Can large language models capture human
  preferences? \emph{Marketing Science} 43(4):709--722.

\bibitem[{Gui \protect\BIBand{} Toubia(2023)}]{gui2023challenge}
Gui G, Toubia O (2023) The challenge of using llms to simulate human behavior:
  A causal inference perspective. \emph{arXiv preprint arXiv:2312.15524} .

\bibitem[{Heckman(1979)}]{heckman1979sample}
Heckman JJ (1979) Sample selection bias as a specification error.
  \emph{Econometrica: Journal of the econometric society} 153--161.

\bibitem[{Horton(2023)}]{horton2023large}
Horton JJ (2023) Large language models as simulated economic agents: What can
  we learn from homo silicus? Technical report, National Bureau of Economic
  Research.

\bibitem[{Hu et~al.(2017)Hu, Pavlou, \protect\BIBand{} Zhang}]{hu2017self}
Hu N, Pavlou PA, Zhang J (2017) On self-selection biases in online product
  reviews. \emph{MIS quarterly} 41(2):449--475.

\bibitem[{Ibrahim et~al.(2001)Ibrahim, Lipsitz, \protect\BIBand{}
  Horton}]{ibrahim2001using}
Ibrahim JG, Lipsitz SR, Horton N (2001) Using auxiliary data for parameter
  estimation with non-ignorably missing outcomes. \emph{Journal of the Royal
  Statistical Society: Series C (Applied Statistics)} 50(3):361--373.

\bibitem[{Imbens \protect\BIBand{} Manski(2004)}]{imbens2004confidence}
Imbens GW, Manski CF (2004) Confidence intervals for partially identified
  parameters. \emph{Econometrica} 72(6):1845--1857.

\bibitem[{Ji et~al.(2025)Ji, Lei, \protect\BIBand{} Zrnic}]{ji2025predictions}
Ji W, Lei L, Zrnic T (2025) Predictions as surrogates: Revisiting surrogate
  outcomes in the age of ai. \emph{arXiv preprint arXiv:2501.09731} .

\bibitem[{Kaido et~al.(2019)Kaido, Molinari, \protect\BIBand{}
  Stoye}]{kaido2019confidence}
Kaido H, Molinari F, Stoye J (2019) Confidence intervals for projections of
  partially identified parameters. \emph{Econometrica} 87(4):1397--1432.

\bibitem[{Kott \protect\BIBand{} Liao(2018)}]{kott2018calibration}
Kott PS, Liao D (2018) Calibration weighting for nonresponse with proxy frame
  variables (so that unit nonresponse can be not missing at random).
  \emph{Journal of Official Statistics} 34(1):107--120.

\bibitem[{Li et~al.(2024)Li, Castelo, Katona, \protect\BIBand{}
  Sarvary}]{li2024frontiers}
Li P, Castelo N, Katona Z, Sarvary M (2024) Frontiers: Determining the validity
  of large language models for automated perceptual analysis. \emph{Marketing
  Science} 43(2):254--266.

\bibitem[{Li et~al.(2025)Li, Wang, \protect\BIBand{} Wang}]{li2025choosing}
Li S, Wang C, Wang J (2025) Choosing the better bandit algorithm under data
  sharing: When do a/b experiments work? \emph{arXiv preprint arXiv:2507.11891}
  .

\bibitem[{Little(1994)}]{little1994class}
Little RJ (1994) A class of pattern-mixture models for normal incomplete data.
  \emph{Biometrika} 81(3):471--483.

\bibitem[{Litvinchev \protect\BIBand{}
  Tsurkov(2013)}]{litvinchev2013aggregation}
Litvinchev I, Tsurkov V (2013) \emph{Aggregation in large-scale optimization},
  volume~83 (Springer Science \& Business Media).

\bibitem[{Lu et~al.(2026)Lu, Wang, Zhang, \protect\BIBand{}
  Zhang}]{lu2026generative}
Lu C, Wang M, Zhang DJ, Zhang H (2026) Generative augmented inference.
  \emph{arXiv preprint arXiv:2604.14575} .

\bibitem[{Manski(2003)}]{manski2003partial}
Manski CF (2003) \emph{Partial identification of probability distributions}
  (Springer).

\bibitem[{Miao et~al.(2024)Miao, Liu, Li, Tchetgen~Tchetgen, \protect\BIBand{}
  Geng}]{miao2024identification}
Miao W, Liu L, Li Y, Tchetgen~Tchetgen EJ, Geng Z (2024) Identification and
  semiparametric efficiency theory of nonignorable missing data with a shadow
  variable. \emph{ACM/JMS Journal of Data Science} 1(2):1--23.

\bibitem[{Miao \protect\BIBand{} Tchetgen~Tchetgen(2016)}]{miao2016varieties}
Miao W, Tchetgen~Tchetgen EJ (2016) On varieties of doubly robust estimators
  under missingness not at random with a shadow variable. \emph{Biometrika}
  103(2):475--482.

\bibitem[{Mogstad et~al.(2018)Mogstad, Santos, \protect\BIBand{}
  Torgovitsky}]{mogstad2018using}
Mogstad M, Santos A, Torgovitsky A (2018) Using instrumental variables for
  inference about policy relevant treatment parameters. \emph{Econometrica}
  86(5):1589--1619.

\bibitem[{Politis \protect\BIBand{} Romano(1994)}]{politis1994large}
Politis DN, Romano JP (1994) Large sample confidence regions based on
  subsamples under minimal assumptions. \emph{The Annals of Statistics}
  22(4):2031--2050.

\bibitem[{Qin et~al.(2008)Qin, Shao, \protect\BIBand{}
  Zhang}]{qin2008efficient}
Qin J, Shao J, Zhang B (2008) Efficient and doubly robust imputation for
  covariate-dependent missing responses. \emph{Journal of the American
  Statistical Association} 103(482):797--810,
  \urlprefix\url{http://dx.doi.org/10.1198/016214508000000238}.

\bibitem[{Rambachan et~al.(2024)Rambachan, Singh, \protect\BIBand{}
  Viviano}]{rambachan2024program}
Rambachan A, Singh R, Viviano D (2024) Program evaluation with remotely sensed
  outcomes. \emph{arXiv preprint arXiv:2411.10959} .

\bibitem[{Robins et~al.(1994)Robins, Rotnitzky, \protect\BIBand{}
  Zhao}]{robins1994estimation}
Robins JM, Rotnitzky A, Zhao LP (1994) Estimation of regression coefficients
  when some regressors are not always observed. \emph{Journal of the American
  Statistical Association} 89(427):846--866,
  \urlprefix\url{http://dx.doi.org/10.1080/01621459.1994.10476818}.

\bibitem[{Rubin(1976)}]{rubin1976inference}
Rubin DB (1976) Inference and missing data. \emph{Biometrika} 63(3):581--592,
  \urlprefix\url{http://dx.doi.org/10.1093/biomet/63.3.581}.

\bibitem[{Rubin(1987)}]{rubin1987calculation}
Rubin DB (1987) The calculation of posterior distributions by data
  augmentation: Comment: A noniterative sampling/importance resampling
  alternative to the data augmentation algorithm for creating a few imputations
  when fractions of missing information are modest: The sir algorithm.
  \emph{Journal of the American Statistical Association} 82(398):543--546.

\bibitem[{Shapiro(1991)}]{shapiro1991asymptotic}
Shapiro A (1991) Asymptotic analysis of stochastic programs. \emph{Annals of
  Operations Research} 30:169--186.

\bibitem[{Sun et~al.(2018)Sun, Liu, Miao, Wirth, Robins, \protect\BIBand{}
  Tchetgen}]{sun2018semiparametric}
Sun B, Liu L, Miao W, Wirth K, Robins J, Tchetgen EJT (2018) Semiparametric
  estimation with data missing not at random using an instrumental variable.
  \emph{Statistica Sinica} 28(4):1965.

\bibitem[{Sun et~al.(2021)Sun, Zhang, Balog, Ren, Ren, Chen, \protect\BIBand{}
  de~Rijke}]{sun2021simulating}
Sun W, Zhang S, Balog K, Ren Z, Ren P, Chen Z, de~Rijke M (2021) Simulating
  user satisfaction for the evaluation of task-oriented dialogue systems.
  \emph{Proceedings of the 44th International ACM SIGIR Conference on Research
  and Development in Information Retrieval}, 2499--2506.

\bibitem[{Tang et~al.(2003)Tang, Little, \protect\BIBand{}
  Raghunathan}]{tang2003analysis}
Tang G, Little RJ, Raghunathan TE (2003) Analysis of multivariate missing data
  with nonignorable nonresponse. \emph{Biometrika} 90(4):747--764.

\bibitem[{Tchetgen~Tchetgen \protect\BIBand{}
  Wirth(2017)}]{tchetgen2017general}
Tchetgen~Tchetgen EJ, Wirth KE (2017) A general instrumental variable framework
  for regression analysis with outcome missing not at random. \emph{Biometrics}
  73(4):1123--1131.

\bibitem[{Toubia et~al.(2025)Toubia, Gui, Peng, Merlau, Li, \protect\BIBand{}
  Chen}]{toubia2025database}
Toubia O, Gui GZ, Peng T, Merlau DJ, Li A, Chen H (2025) Database report:
  Twin-2k-500: A data set for building digital twins of over 2,000 people based
  on their answers to over 500 questions. \emph{Marketing Science}
  44(6):1446--1455.

\bibitem[{Voronin(2025)}]{voronin2025linear}
Voronin A (2025) Linear programming approach to partially identified
  econometric models. \emph{arXiv preprint arXiv:2503.14940} .

\bibitem[{Wang et~al.(2025)Wang, Ye, \protect\BIBand{}
  Zhao}]{wang2025efficient}
Wang L, Ye Z, Zhao J (2025) Efficient inference using large language models
  with limited human data: Fine-tuning then rectification. \emph{arXiv preprint
  arXiv:2511.19486} .

\bibitem[{Wang et~al.(2024)Wang, Zhang, \protect\BIBand{}
  Zhang}]{wang2024large}
Wang M, Zhang DJ, Zhang H (2024) Large language models for market research: A
  data-augmentation approach. \emph{arXiv preprint arXiv:2412.19363} .

\bibitem[{Xu et~al.(2026)Xu, Liang, Liu, Mao, \protect\BIBand{}
  Wu}]{xu2026uncertainty}
Xu J, Liang M, Liu G, Mao X, Wu J (2026) Uncertainty-guided llm semantic
  augmentation for heterogeneous treatment effect estimation. \emph{arXiv
  preprint arXiv:2607.26599} .

\bibitem[{Ye et~al.(2025)Ye, Yoganarasimhan, \protect\BIBand{}
  Zheng}]{ye2025lola}
Ye Z, Yoganarasimhan H, Zheng Y (2025) Lola: Llm-assisted online learning
  algorithm for content experiments. \emph{Marketing Science} 44(5):995--1016.

\bibitem[{Zhang et~al.(2026)Zhang, Han, Hu, \protect\BIBand{}
  Peng}]{zhang2026llm}
Zhang Z, Han J, Hu M, Peng Y (2026) Llm-inspired pretrain-then-finetune for
  small-data, large-scale optimization. \emph{arXiv preprint arXiv:2602.03690}
  .

\bibitem[{Zipkin(1980)}]{zipkin1980bounds}
Zipkin PH (1980) Bounds for row-aggregation in linear programming.
  \emph{Operations Research} 28(4):903--916.

\end{thebibliography}
}

\newpage
\setcounter{page}{1}
\begin{APPENDICES}
\section{Connection to the Shadow Variable Framework}\label{apx.sec:connection}

Assumption~\ref{ass:cond-indep} is closely related to the \textit{shadow variable} or \textit{auxiliary variable} framework, which has been proposed as an alternative to instrumental-variable approaches for nonrandom missing data \citep{d2010new, miao2016varieties, miao2024identification} and has been used in empirical studies \citep{ibrahim2001using, kott2018calibration}. In the classical framework, a shadow variable is fully observed, relevant for the outcome, and excluded from the response mechanism conditional on $(Y,X)$, as in Assumption~\ref{ass:cond-indep}. Point identification further requires a \emph{completeness} condition on the conditional distribution $\P(Y \mid R=1, X, F)$ \citep{miao2016varieties, miao2024identification}.

\begin{definition}[Completeness of $\P(Y \mid X, F, R=1)$]\label{def:completeness}
For a shadow variable $F$, the conditional distribution $\P(Y \mid X, F, R=1)$ is called complete if, for each $x$ and every square-integrable function $h(x,Y)$, $\E[h(x,Y) \mid X=x, F, R=1] = 0$ almost surely implies $h(x,Y) = 0$ almost surely.
\end{definition}

The completeness condition strengthens ordinary relevance: not only must the shadow variable $F$ be associated with the outcome $Y$, but the variation in $F$ must be rich enough to recover variation in $Y$. Under our discrete outcome setup, the condition has a simple matrix interpretation. For a fixed covariate value $x$, define the joint distribution matrix $H_x = [P(F=f, Y=y \mid X=x)]_{f \in \mathcal{F}, y \in [M]} \in \mathbb{R}^{|\mathcal{F}| \times M}$ and the respondent conditional distribution matrix $B_x = [P(Y=y \mid F=f, X=x, R=1)]_{f \in \mathcal{F}, y \in [M]} \in \mathbb{R}^{|\mathcal{F}| \times M}$. The matrix $H_x$ represents the full-data association between $F$ and $Y$ conditional on $X=x$, whereas $B_x$ describes the corresponding association among observed respondents. Completeness corresponds to a full-rank condition on $B_x$, which is equivalent to a rank condition on $H_x$ under mild positivity assumptions.

\begin{proposition}\label{prop:completeness-equivalence}
    For discrete outcomes $Y$ and a shadow variable $F$, fix any covariate value $x$. The completeness condition at $x$ holds if and only if $\operatorname{rank}(B_x) = M$. Furthermore, if $\pi_x(y)>0$ and $P(F=f, R=1\mid X=x) >0$ for all $x, y$ and $f$, then the completeness condition at $x$ is equivalent to $\operatorname{rank}(H_x) = M$.
\end{proposition}

Proposition~\ref{prop:completeness-equivalence} translates the abstract completeness condition into a concrete matrix rank condition. In particular, completeness requires the matrix $H_x$ to have full column rank, which can fail if the shadow variable has smaller support than the outcome. For example, a binary indicator of task completion cannot satisfy full-rank completeness for a five-point rating outcome. Even when $F$ and $Y$ have the same support size, completeness requires the distribution of $F$ to vary in linearly independent ways across outcome values. In modern applications where predictive models are generic rather than problem-specific---as is often the case with large language models---the predictions may not fully capture variation in the outcome, leading to ill-posed point identification.



Our definition of a weak shadow variable departs from the classical shadow-variable literature in an important way: we fully relax completeness, which allows us to incorporate lower-dimensional auxiliary information. Completeness is not necessary to identify bounds on $\theta$. Instead, the proposed bounds adapt to the strength of association between $Y$ and $F$ conditional on $X$ and $R=1$. Stronger association leads to tighter identification regions, and completeness recovers point identification as a limiting case. This perspective allows practitioners to use modern predictive models as auxiliary tools for tightening identification regions without imposing stringent requirements on predictive accuracy.

\section{Additional Results for Experiment}
\subsection{Data Examples}\label{app:data-examples}

Below are two example dialogues from the JDDC dataset (translated from Chinese), illustrating contrasting customer experiences. Each user turn includes satisfaction scores from three independent annotators on a 1--5 scale.

\paragraph{Example 1: Satisfied Customer (Median Rating = 4)}
This dialogue shows a customer canceling multiple orders, with the system successfully processing the requests.

\begin{small}
\begin{verbatim}
USER    Are you there?                                    3,3,3
USER    I placed several orders today                     3,3,3
SYSTEM  What problem can I help you with?
USER    And I need them urgently                          3,3,3
USER    So I bought from a physical store instead         3,3,3
SYSTEM  Order [ORDERID_10294471] has been canceled
USER    Can I get a refund?                               3,3,3
SYSTEM  Yes, you can
USER    What about the other three orders?                3,3,3
...
USER    OK, thank you                                     4,4,4
SYSTEM  Is there anything else I can help you with?
USER    OVERALL                                           4,5,3
\end{verbatim}
\end{small}

\noindent Annotator explanations: ``System cannot solve the problem, needs merchant handling''; ``System resolved the customer's issue well''; ``System understood the need but cannot resolve, transferred to merchant.''

\paragraph{Example 2: Dissatisfied Customer (Median Rating = 1)}
This dialogue shows a customer seeking information about a service center location, with the system failing to understand and assist.

\begin{small}
\begin{verbatim}
USER    I want to exchange goods                          3,3,3
USER    I don't know where the service center is          3,3,3
SYSTEM  Wait, let me check
...
USER    Can you help me faster?                           2,2,3
USER    So slow                                           1,1,2
USER    I just want to find the service center            3,2,3
USER    How long? I can't keep watching my phone          1,1,2
SYSTEM  Hello, you can return and buy again
USER    Are you crazy?                                    1,1,1
...
USER    If you don't understand, get someone who does     1,1,1
USER    I don't have time to waste                        1,1,3
SYSTEM  We really can't find the service center
USER    OVERALL                                           1,1,1
\end{verbatim}
\end{small}

\noindent Annotator explanations: ``System cannot solve problem, doesn't understand user''; ``System completely failed to solve problem or understand customer intent, very poor experience''; ``System did not understand user's needs, poor experience.''

\subsection{Definition for Relevance and Exclusion}\label{app:prompt-diagnostics}

This section defines the prompt-level diagnostics used in Figure~\ref{fig:prompt-relevance-independence}. We compute each diagnostic separately for every valid combination of prompt family, prompt-side LLM, and missingness-side LLM. Let $i=1,\ldots,n_c$ index the dialogues available for one such combination $c$, and let $(F_i,Y_i,R_i,p_i)$ denote the extracted shadow score, the observed full-data rating, the realized response indicator, and the simulated response propensity, respectively. In the semi-synthetic design, $R_i$ is the single response realization returned by the missingness-side LLM together with $p_i$.

\paragraph{Relevance.}
For each model combination $c$, we measure relevance using the sample Spearman correlation
\[
\widehat{\rho}_{S,c}
=
\operatorname{Corr}\bigl(\operatorname{rank}(F_i),
\operatorname{rank}(Y_i)\bigr),
\]
where tied values receive their average ranks. This rank-based measure accommodates both the ordinal $1$--$5$ scores and the binary prompt outputs. The relevance coordinate reported for a prompt is the average of $\widehat{\rho}_{S,c}$ over its valid prompt-side and missingness-side LLM combinations.

\paragraph{Exclusion test.}
To assess the exclusion restriction $F\indep R\mid Y$, we form a two-way contingency table of $(F,R)$ within each outcome stratum. Let
\[
N_{yfr}=\sum_{i=1}^{n_c}\mathds{1}\{Y_i=y,F_i=f,R_i=r\},
\qquad
E_{yfr}=\frac{N_{yf+}N_{y+r}}{N_{y++}},
\]
where $E_{yfr}$ is the expected cell count under the exclusion restriction. We retain the informative outcome strata in which both $F$ and $R$ take at least two distinct values and compute the stratified likelihood-ratio statistic
\[
G_c^2
=
2\sum_{y}\sum_{f,r:N_{yfr}>0}
N_{yfr}\log\left(\frac{N_{yfr}}{E_{yfr}}\right).
\]
Under the exclusion null, we compare $G_c^2$ with a chi-square distribution having
\[
d_c=\sum_y (|\mathcal F_{c,y}|-1)(|\mathcal R_{c,y}|-1)
\]
degrees of freedom, where the sum is over informative strata and $\mathcal F_{c,y}$ and $\mathcal R_{c,y}$ are the observed supports within stratum $y$. The test does not reject the exclusion restriction for a model combination when its $p$-value exceeds $0.05$. The left panel of Figure~\ref{fig:prompt-relevance-independence} reports, for each prompt, the fraction of valid model combinations for which the test does not reject. Thus a higher non-rejection rate indicates weaker statistical evidence against Assumption~\ref{ass:cond-indep}; it does not by itself establish that the exclusion restriction holds exactly.

\paragraph{Response-propensity gap.}
Because the semi-synthetic design also records the response propensity $p_i$, we supplement the test with a magnitude-based diagnostic. For each informative outcome stratum $y$, define
\[
\widehat p_{c}(f,y)
=
\frac{\sum_{i=1}^{n_c}p_i\mathds{1}\{F_i=f,Y_i=y\}}
{\sum_{i=1}^{n_c}\mathds{1}\{F_i=f,Y_i=y\}}
\]
and
\[
\widehat\Delta_c
=
\frac{1}{|\mathcal Y_c^{\mathrm{inf}}|}
\sum_{y\in\mathcal Y_c^{\mathrm{inf}}}
\left\{
\max_{f\in\mathcal F_{c,y}}\widehat p_c(f,y)
-
\min_{f\in\mathcal F_{c,y}}\widehat p_c(f,y)
\right\},
\]
where $\mathcal Y_c^{\mathrm{inf}}$ contains the outcome strata with at least two observed values of $F$. The prompt-level gap in the right panel of Figure~\ref{fig:prompt-relevance-independence} averages $\widehat\Delta_c$ over valid model combinations. When the exclusion restriction holds exactly, the population response propensity does not vary with $f$ after conditioning on $Y$, so the corresponding population gap is zero. Smaller values of $\widehat\Delta_c$ therefore indicate smaller departures from the weak-shadow-variable exclusion restriction.

\subsection{Robustness Across Language Models}\label{app:model-robustness}

We assess whether the empirical conclusions depend on the choice of language model by separating the two roles that models play in the experiment. The prompt-side model produces the candidate shadow variable $F$, whereas the missingness-side model acts as a digital twin and generates the response propensity. Panel A of Table~\ref{tab:model-robustness} averages each prompt-side model over the 12 prompt families and three missingness-side models; Panel B reverses this aggregation. For the downstream outcomes, we first average each structural cell over 100 Bernoulli response resamples. The non-rejection rate and propensity gap follow Appendix~\ref{app:prompt-diagnostics}.

\begin{table}[htbp]
    \centering
    \small
    \caption{Robustness across prompt-side and missingness-side language models. Midpoint MAE and interval width refer to the localized estimator. A higher non-rejection rate and a smaller propensity gap indicate closer agreement with the exclusion restriction.}
    \label{tab:model-robustness}
    \setlength{\tabcolsep}{4.2pt}
    \begin{tabular}{lccccc}
        \toprule
        Model & $\mathrm{corr}(F,Y)$ & Non-rejection & Propensity gap & Midpoint MAE & Interval width \\
        \midrule
        \multicolumn{6}{l}{\textit{Panel A: Prompt-side model generating $F$}} \\
        GPT-4.1-mini         & 0.323 & 0.750 & 0.0326 & 0.0561 & 0.1699 \\
        GPT-5.4-mini         & 0.349 & 0.722 & 0.0276 & 0.0582 & 0.2811 \\
        Grok-4-20-Reasoning  & 0.325 & 0.778 & 0.0327 & 0.0644 & 0.2457 \\
        \addlinespace
        \multicolumn{6}{l}{\textit{Panel B: Missingness-side model generating response propensities}} \\
        DeepSeek-V3.2        & ---   & 0.833 & 0.0462 & 0.0584 & 0.2333 \\
        GPT-5.4-mini         & ---   & 0.528 & 0.0173 & 0.0628 & 0.2516 \\
        Grok-4-20-Reasoning  & ---   & 0.889 & 0.0294 & 0.0575 & 0.2117 \\
        \bottomrule
    \end{tabular}
\end{table}

Panel A shows that the prompt-side models differ more in downstream interval width than in relevance. GPT-5.4-mini delivers the highest rank correlation with the outcome, at $0.349$, and Grok-4-20-Reasoning has the highest non-rejection rate, at $0.778$. GPT-4.1-mini nevertheless performs best downstream, with midpoint MAE $0.0561$ and interval width $0.1699$. Thus scalar relevance alone does not determine the identifying content of $F$.

Panel B indicates greater sensitivity to the missingness-side model. Grok-4-20-Reasoning produces the highest non-rejection rate and the narrowest downstream interval, while GPT-5.4-mini produces the smallest propensity gap but a lower non-rejection rate. DeepSeek-V3.2 gives similar midpoint accuracy but the largest propensity gap. The test and gap are therefore complementary: one measures statistical evidence using the realized response indicator, while the other measures the magnitude of dependence using simulated probabilities. Because these models define different semi-synthetic data-generating processes, the comparison is a robustness exercise rather than a test of predictive accuracy against observed customer behavior.

Most importantly, the estimator comparison is stable across all nine model pairings. The localized midpoint MAE ranges from $0.053$ to $0.068$, which is $26\%$--$49\%$ below the best competing point estimator within each pairing. Its interval width ranges from $0.158$ to $0.306$, compared with $2.02$--$2.07$ for aggregated LP, an $85\%$--$92\%$ reduction. Thus model choice affects the relevance--exclusion frontier and the degree of interval tightening, but it does not overturn the central finding that LLM-generated weak shadow variables improve identification and estimation under MNAR outcomes.

\section{Proofs}

\subsection{Proofs for Identification Results}\label{apx:mean-proofs}

\subsubsection{Proof of Proposition~\ref{prop:no-shadow-bound}}

The only information about the missing outcomes is their total probability mass. It is therefore useful to separate the observed mass from the missing mass. Let
\[
p(y)=\alpha(y)\{1+w(y)\},\qquad q(y):=p(y)-\alpha(y).
\]
The constraints in \eqref{opt:mean-mnar-range} are equivalent to $q(y)\ge0$ and
\[
\sum_{y=1}^M q(y)=1-\sum_{y=1}^M\alpha(y)=\P(R=0).
\]
The objective can be written as
\[
\sum_{y=1}^M y\,p(y)
=
\sum_{y=1}^M y\,\alpha(y)+\sum_{y=1}^M y\,q(y).
\]
Thus the problem is to allocate the unidentified missing mass $\P(R=0)$ across the support $\{1,\ldots,M\}$.

The lower endpoint is obtained by assigning all missing mass to the lowest rating, so
\[
\theta_{\min}
=
\sum_{y=1}^M y\,\alpha(y)+\P(R=0)
=
\P(R=1)\E[Y\mid R=1]+\P(R=0).
\]
Similarly, the upper endpoint is obtained by assigning all missing mass to the highest rating:
\[
\theta_{\max}
=
\sum_{y=1}^M y\,\alpha(y)+M\P(R=0)
=
\P(R=1)\E[Y\mid R=1]+M\P(R=0).
\]
These allocations also show sharpness, because every value between the two endpoints is obtained by redistributing the same missing mass between lower and higher outcome levels.

\subsubsection{Proof of Proposition~\ref{prop:completeness-equivalence}}

Fix a covariate value $x$. For discrete $Y \in [M]$, the completeness condition in Definition~\ref{def:completeness} reduces to: for any $h_x: [M] \to \mathbb{R}$,
\[
\E[h_x(Y) \mid X=x, F = f, R = 1] = 0 \quad \text{for all } f \in \mathcal{F} \quad \Longrightarrow \quad h_x(y) = 0 \quad \text{for all } y \in [M].
\]
Writing $(B_x)_{fy} = P(Y = y \mid X=x, F = f, R = 1)$, the left-hand condition is $\sum_{y=1}^M (B_x)_{fy}\, h_x(y) = 0$ for all $f$, i.e., $B_x\mathbf{h}_x = \bm{0}$. Hence completeness at $x$ holds if and only if the null space of $B_x$ is trivial, which is equivalent to $\operatorname{rank}(B_x) = M$.

Next, we relate $B_x$ to $H_x$. Under Assumption~\ref{ass:cond-indep}, $P(R = 1 \mid Y = y, F = f, X=x) = \pi_x(y)$, so
\[
(B_x)_{fy} = \frac{P(Y = y, F = f, R = 1\mid X=x)}{P(F = f, R = 1\mid X=x)} = \frac{\pi_x(y)\, (H_x)_{fy}}{P(F = f, R = 1\mid X=x)}.
\]
In matrix form, $B_x = D_{p,x}^{-1} H_x D_{\pi,x}$, where $D_{p,x} = \operatorname{diag}(P(F = f, R = 1\mid X=x))_{f \in \mathcal{F}}$ and $D_{\pi,x} = \operatorname{diag}(\pi_x(1), \ldots, \pi_x(M))$. Under the stated positivity conditions, both $D_{p,x}$ and $D_{\pi,x}$ are invertible, so $\operatorname{rank}(B_x) = \operatorname{rank}(H_x)$, and the completeness condition at $x$ is equivalent to $\operatorname{rank}(H_x) = M$.

\subsubsection{Proof of Theorem~\ref{thm:point-id-mean}}

Fix a covariate value $x$. The observed law identifies $A_x$ and $\bm\beta_x$, while the only remaining unknowns are the nonresponse odds $\mathbf w_x$. Under Assumption~\ref{ass:cond-indep}, the accounting identity for the missing observations is exactly
\[
A_x\mathbf w_x=\bm\beta_x,\qquad \mathbf w_x\ge0.
\]
For any full-data law satisfying the assumption, its odds vector therefore belongs to this feasible set, and its conditional mean equals the objective in
\[
\theta_{x,\min/\max} = \min/\max_{\mathbf{w}_x} \bm{1}^\top A_x D (\mathbf{w}_x + \bm{1}) \quad \text{s.t.} \quad A_x \mathbf{w}_x = \bm{\beta}_x, \quad \mathbf{w}_x \geq \bm{0}.
\]
Conversely, any feasible $\mathbf w_x$ defines a full-data conditional distribution by setting
\[
\P(Y=y,F=f\mid X=x)=\alpha_x(f,y)\{1+w_x(y)\},
\qquad
\P(R=1\mid Y=y,X=x)=\frac{1}{1+w_x(y)}.
\]
Then $\P(R=1,F=f,Y=y\mid X=x)=\alpha_x(f,y)$ and $\P(R=0,F=f,Y=y\mid X=x)=\alpha_x(f,y)w_x(y)$. The constraint $A_x\mathbf w_x=\bm\beta_x$ ensures that the constructed law reproduces both the observed respondent distribution and the observed nonrespondent distribution, and the response probability depends on $(Y,X)$ but not on $F$. Hence the LP endpoints form the sharp conditional identification interval for $\theta_x$. Moreover, because the feasible set is convex and the objective is linear, every value between $\theta_{x,\min}$ and $\theta_{x,\max}$ is attained by a convex combination of conditional feasible odds vectors.

Finally, $\theta=\E_X[\theta_X]$. The pointwise constructions can be combined across covariate values because each construction matches the same conditional observed law given $X=x$ and leaves the marginal law of $X$ unchanged. Choosing conditional minimizers for $P_X$-almost every $x$ attains $\E_X[\theta_{X,\min}]$, and choosing conditional maximizers attains $\E_X[\theta_{X,\max}]$. Convexly mixing the corresponding feasible odds functions pointwise in $x$ attains every intermediate value between these two aggregate endpoints. Therefore averaging the conditional sharp intervals over the marginal distribution of $X$ gives the sharp interval stated in the theorem. If $A_x$ has full column rank for $P_X$-almost every $x$, then $A_x\mathbf w_x=\bm\beta_x$ has at most one solution. Feasibility gives exactly one solution, so $\theta_{x,\min}=\theta_{x,\max}$ for $P_X$-almost every $x$, and $\theta$ is point identified.

\subsubsection{Proof of Proposition~\ref{prop:lp-bound}}\label{apx:prop-lp-bound-proof}

We compare the shadow-variable LP to the no-shadow LP after conditioning on a covariate value. The comparison holds for $P_X$-almost every $x$, and the unconditional inequalities follow by taking expectations with respect to the marginal law of $X$.

Fix such an $x$ and write $A_x=[\bm a_{x,1},\ldots,\bm a_{x,M}]$. Let
\[
s_{x,y}:=\bm1^\top\bm a_{x,y},\qquad
b_x:=\bm1^\top\bm\beta_x.
\]
If $b_x=0$, there is no missing mass at this value of $x$, and the contribution of this $x$ to both sides of the desired inequalities is zero. Hence consider $b_x>0$. Define the shadow feasible set and its aggregated relaxation as
\[
\mathcal W_x:=\{\mathbf w\ge\bm0:A_x\mathbf w=\bm\beta_x\},
\qquad
\widetilde{\mathcal W}_x:=\{\mathbf w\ge\bm0:\bm1^\top A_x\mathbf w=b_x\}.
\]
The aggregated set is exactly the conditional no-shadow relaxation: it retains only the total missing mass constraint and drops the distributional restrictions across values of $F$. Since $A_x\mathbf w=\bm\beta_x$ implies $\bm1^\top A_x\mathbf w=b_x$, we have $\mathcal W_x\subseteq\widetilde{\mathcal W}_x$. Therefore the shadow upper endpoint cannot exceed the no-shadow upper endpoint, and the shadow lower endpoint cannot fall below the no-shadow lower endpoint.

Formally, define the conditional aggregated endpoints by replacing $\mathcal W_x$ with $\widetilde{\mathcal W}_x$ in the LP objective:
\[
\widetilde\theta_{x,\max}:=\max_{\mathbf w\in\widetilde{\mathcal W}_x}\bm1^\top A_xD(\mathbf w+\bm1),
\qquad
\widetilde\theta_{x,\min}:=\min_{\mathbf w\in\widetilde{\mathcal W}_x}\bm1^\top A_xD(\mathbf w+\bm1).
\]

The rest of the proof quantifies this containment. For any $\mathbf w\in\mathcal W_x$, define
\[
\lambda_{x,y}:=\frac{s_{x,y}w(y)}{b_x}.
\]
These weights are nonnegative and sum to one because
\[
\sum_{y=1}^M\lambda_{x,y}
=
\frac{\bm1^\top A_x\mathbf w}{b_x}
=1.
\]
For columns with $s_{x,y}>0$, let $\bm p_{x,y}:=\bm a_{x,y}/s_{x,y}$. Columns with $s_{x,y}=0$ have $\lambda_{x,y}=0$ and do not affect the following convex combination. Dividing the equality $A_x\mathbf w=\bm\beta_x$ by $b_x$ gives
\[
\frac{\bm\beta_x}{b_x}
=
\sum_{y:s_{x,y}>0}\lambda_{x,y}\bm p_{x,y}.
\]
Thus the normalized missing distribution over $F$ is a convex combination of the normalized observed columns. If $s_{x,M}>0$, then
\[
\left\|\frac{\bm\beta_x}{b_x}-\frac{\bm a_{x,M}}{s_{x,M}}\right\|_1
=
\left\|\sum_{\substack{y\neq M:\\ s_{x,y}>0}}\lambda_{x,y}(\bm p_{x,y}-\bm p_{x,M})\right\|_1
\le
2(1-\lambda_{x,M}),
\]
because each $\bm p_{x,y}$ is a probability vector. Similarly, if $s_{x,1}>0$, then
\[
\left\|\frac{\bm\beta_x}{b_x}-\frac{\bm a_{x,1}}{s_{x,1}}\right\|_1
\le
2(1-\lambda_{x,1}).
\]

We next translate these mass-composition bounds into objective gaps. The constant observed component $\bm1^\top A_xD\bm1$ is common to the shadow and no-shadow objectives, so it cancels from the comparison. The missing component satisfies
\[
\bm1^\top A_xD\mathbf w
=
\sum_{y=1}^M y\,s_{x,y}w(y)
=
b_x\sum_{y=1}^M y\,\lambda_{x,y}.
\]
If $s_{x,M}>0$, the no-shadow relaxation can place all missing mass on outcome $M$ by choosing $w(M)=b_x/s_{x,M}$ and $w(y)=0$ for $y\neq M$, and no feasible allocation can exceed value $b_xM$. Let $\mathbf w_x^\star$ be an upper-endpoint optimizer of the shadow LP, with associated weights $\lambda_{x,y}^\star$. Then
\[
\widetilde\theta_{x,\max}-\theta_{x,\max}^{\mathrm{shad}}
=
b_xM-b_x\sum_{y=1}^M y\,\lambda_{x,y}^\star
\ge
b_x(1-\lambda_{x,M}^\star)
\ge
\frac{b_x}{2}
\left\|\frac{\bm\beta_x}{b_x}-\frac{\bm a_{x,M}}{s_{x,M}}\right\|_1.
\]
If $s_{x,M}=0$, the normalized term in Proposition~\ref{prop:lp-bound} is defined to contribute zero, and the same nonnegativity follows from $\mathcal W_x\subseteq\widetilde{\mathcal W}_x$.

The lower endpoint is analogous. If $s_{x,1}>0$, the no-shadow relaxation can place all missing mass on outcome $1$ by choosing $w(1)=b_x/s_{x,1}$ and $w(y)=0$ for $y\neq1$, and no feasible allocation can have missing component below $b_x$. Let $\mathbf w_{x,\star}$ be a lower-endpoint optimizer of the shadow LP, with associated weights $\lambda_{x,y,\star}$. Then
\[
\theta_{x,\min}^{\mathrm{shad}}-\widetilde\theta_{x,\min}
=
b_x\sum_{y=1}^M y\,\lambda_{x,y,\star}-b_x
\ge
b_x(1-\lambda_{x,1,\star})
\ge
\frac{b_x}{2}
\left\|\frac{\bm\beta_x}{b_x}-\frac{\bm a_{x,1}}{s_{x,1}}\right\|_1.
\]
If $s_{x,1}=0$, the corresponding normalized term is again interpreted as zero, and the containment of feasible sets gives the required nonnegative gap.

Finally, the unconditional no-shadow endpoints are the expectations of the conditional aggregated endpoints, while Theorem~\ref{thm:point-id-mean} gives
\[
\theta_{\max}^{\mathrm{shad}}=\E_X[\theta_{X,\max}^{\mathrm{shad}}],
\qquad
\theta_{\min}^{\mathrm{shad}}=\E_X[\theta_{X,\min}^{\mathrm{shad}}].
\]
Taking expectations of the pointwise bounds over $X$ yields
\[
\theta_{\max}-\theta_{\max}^{\mathrm{shad}}
\ge
\E_X\left[
\frac{\bm1^\top\bm\beta_X}{2}
\left\|
\frac{\bm\beta_X}{\bm1^\top\bm\beta_X}
-
\frac{\bm a_{X,M}}{\bm1^\top\bm a_{X,M}}
\right\|_1
\right]\ge0,
\]
and
\[
\theta_{\min}^{\mathrm{shad}}-\theta_{\min}
\ge
\E_X\left[
\frac{\bm1^\top\bm\beta_X}{2}
\left\|
\frac{\bm\beta_X}{\bm1^\top\bm\beta_X}
-
\frac{\bm a_{X,1}}{\bm1^\top\bm a_{X,1}}
\right\|_1
\right]\ge0.
\]
This proves the two unconditional inequalities for a general covariate distribution.

\subsubsection{Proof of Theorem~\ref{thm:shadow-sensitivity-lp}}
Fix $x\in\mathcal X$. For any full-data law satisfying Assumption~\ref{ass:shadow-sensitivity}, the accounting identities in \eqref{opt:shadow-sensitivity-lp} hold by construction, and the odds-ratio restriction in Assumption~\ref{ass:shadow-sensitivity} gives the third group of constraints. Hence every admissible law induces a feasible point of \eqref{opt:shadow-sensitivity-lp} with objective value equal to $\theta_x$.

Conversely, let $(w_x,u_x)$ satisfy \eqref{opt:shadow-sensitivity-lp}. Define
\[
p_x(y):=\alpha_x(y)\{1+w_x(y)\},\qquad y\in[M].
\]
Using the second group of constraints and then the first,
\[
\sum_{y=1}^M p_x(y)
=
\sum_{y=1}^M\alpha_x(y)+\sum_{y=1}^M\sum_{f\in\mathcal F}\alpha_x(f,y)u_x(f,y)
=
\P(R=1\mid X=x)+\sum_{f\in\mathcal F}\beta_x(f)
=1.
\]
Hence $p_x(\cdot)$ is a probability mass function. For every $y$ with $p_x(y)>0$, define
\[
\P(Y=y\mid X=x)=p_x(y),
\]
\[
\P(R=1,F=f\mid Y=y,X=x)=\frac{\alpha_x(f,y)}{p_x(y)},
\qquad
\P(R=0,F=f\mid Y=y,X=x)=\frac{\alpha_x(f,y)u_x(f,y)}{p_x(y)}.
\]
For $y$ with $p_x(y)=0$, define the conditional law arbitrarily. The column constraints imply that the conditional probabilities above sum to one for each $y$, while the row constraints reproduce $\beta_x(f)$. By construction,
\[
\P(R=1,F=f,Y=y\mid X=x)=\alpha_x(f,y),
\]
and
\[
\P(R=0,F=f\mid X=x)
=
\sum_{y=1}^M \alpha_x(f,y)u_x(f,y)
=\beta_x(f),
\]
so the observed law is matched exactly. The same construction also recovers the intended odds variables. For every $y$ with $\alpha_x(y)>0$,
\[
\frac{\P(R=0,Y=y\mid X=x)}{\P(R=1,Y=y\mid X=x)}
=
\frac{\sum_{f\in\mathcal F}\alpha_x(f,y)u_x(f,y)}{\sum_{f\in\mathcal F}\alpha_x(f,y)}
=
w_x(y),
\]
where the last equality uses the column constraint in \eqref{opt:shadow-sensitivity-lp}. Similarly, for every cell with $\alpha_x(f,y)>0$,
\[
\frac{\P(R=0,F=f,Y=y\mid X=x)}{\P(R=1,F=f,Y=y\mid X=x)}
=u_x(f,y).
\]
Thus the third group of constraints is exactly the odds-ratio band in Assumption~\ref{ass:shadow-sensitivity}. Finally, the objective in \eqref{opt:shadow-sensitivity-lp} equals $\theta_x$. This proves sharpness for $P_X$-almost every covariate value $x$, and taking expectations over $X$ yields the stated interval for $\theta$.

\subsection{Proofs for Estimation and Inference Results}\label{apx:bootstrap-proofs}

We prove the claims for the lower endpoint. The upper endpoint follows by replacing the lower-endpoint objective loading with its negative. Write
\[
a(A):=\bm1^\top A D\bm1,
\qquad
\bm c(A):=D A^\top\bm1,
\]
so that $B_K(\eta)=a(A)+\min_{0\le\mathbf w\le K\bm1}\{\bm c(A)^\top\mathbf w+K\|A\mathbf w-\bm\beta\|_1\}$. Fix any norm $\|\cdot\|$ on the finite-dimensional parameter space of $\eta=(A,\bm\beta)$.

\subsubsection{Proof of Lemma~\ref{lem:Gamma-zero-main}}

Fix any feasible $(\mathbf w,\bm\lambda)$ in \eqref{eq:localized-gap}. Write
\[
\mathbf r:=A\mathbf w-\bm\beta,
\qquad
\mathbf u:=[A^\top\bm\lambda-\bm c(A)]_+.
\]
Since $\bm c(A)-A^\top\bm\lambda\ge-\mathbf u$ componentwise and $\mathbf w\ge0$,
\[
\mathbf w^\top\{\bm c(A)-A^\top\bm\lambda\}\ge-\mathbf w^\top\mathbf u.
\]
Also, $\bm\lambda^\top\mathbf r\ge-\|\bm\lambda\|_\infty\|\mathbf r\|_1$. Therefore,
\begin{align*}
&\bm c(A)^\top\mathbf w-\bm\beta^\top\bm\lambda
+2K\|\mathbf r\|_1
+2K\|\mathbf u\|_1 \\
&\qquad=
\mathbf w^\top\{\bm c(A)-A^\top\bm\lambda\}
+\bm\lambda^\top\mathbf r
+2K\|\mathbf r\|_1
+2K\|\mathbf u\|_1 \\
&\qquad\ge
-\mathbf w^\top\mathbf u-\|\bm\lambda\|_\infty\|\mathbf r\|_1
+2K\|\mathbf r\|_1
+2K\|\mathbf u\|_1 \\
&\qquad=
(2K\bm1-\mathbf w)^\top\mathbf u
+(2K-\|\bm\lambda\|_\infty)\|\mathbf r\|_1
\ge
K\|\mathbf u\|_1+K\|\mathbf r\|_1\ge0.
\end{align*}
Taking the infimum proves $\Gamma_K(\eta)\ge0$.

If an optimal pair lies in the box, then $A_0\mathbf w_0^\star=\bm\beta_0$, $A_0^\top\bm\lambda_0^\star\le\bm c(A_0)$, and the primal-dual gap is zero. Plugging this pair into \eqref{eq:localized-gap} gives $\Gamma_K(\eta_0)=0$. Conversely, if $\Gamma_K(\eta_0)=0$, compactness of the box implies that the minimum is attained. The lower bound above then forces the residual and positive dual violation to be zero, and the remaining objective is the primal-dual gap. Thus the attaining pair is a population primal-dual optimal pair in the box.

\subsubsection{Proof of Proposition~\ref{prop:fix-penalty}}

We first record the fixed-radius dual form used in the argument. Linearize the $\ell_1$ norm in \eqref{eq:localized-lower-bound} by introducing $\mathbf r^+,\mathbf r^-\ge0$ with
\[
A\mathbf w+\mathbf r^+-\mathbf r^-=\bm\beta.
\]
Then $B_K(\eta)-a(A)$ is the value of
\[
\min_{\mathbf w,\mathbf r^+,\mathbf r^-}
\Bigl\{\bm c(A)^\top\mathbf w+K\bm 1^\top(\mathbf r^++\mathbf r^-):
A\mathbf w+\mathbf r^+-\mathbf r^-=\bm\beta,\quad
0\le \mathbf w\le K\bm 1,\quad
\mathbf r^+,\mathbf r^-\ge0\Bigr\}.
\]
Replacing the upper bound $\mathbf w\le K\bm1$ by a slack $\mathbf q\ge0$ satisfying $\mathbf w+\mathbf q=K\bm1$, the dual problem is
\[
\max_{\bm\lambda,\mathbf u\ge0}
\Bigl\{\bm\beta^\top\bm\lambda-K\bm1^\top\mathbf u:
A^\top\bm\lambda-\mathbf u\le \bm c(A),\quad
-K\bm1\le \bm\lambda\le K\bm1\Bigr\}.
\]
Strong duality applies because the primal problem is feasible and bounded for every $\eta$. For fixed $\bm\lambda$, the optimal choice is $\mathbf u=[A^\top\bm\lambda-\bm c(A)]_+$, which yields
\begin{equation}\label{eq:apx-BK-dual}
B_K(\eta)
=
a(A)
+
\max_{\|\bm\lambda\|_\infty\le K}
\Bigl\{\bm\beta^\top\bm\lambda-K\bigl\|\bigl[A^\top\bm\lambda-\bm c(A)\bigr]_+\bigr\|_1\Bigr\}.
\end{equation}

Now suppose the population optimal pair $(\mathbf w_0^\star,\bm\lambda_0^\star)$ lies in the radius-$K$ box. Since $A_0^\top\bm\lambda_0^\star\le\bm c(A_0)$, the penalty term in \eqref{eq:apx-BK-dual} vanishes at $\bm\lambda_0^\star$, and hence
\[
B_K(\eta_0)
\ge
a(A_0)+\bm\beta_0^\top\bm\lambda_0^\star
=
\theta_{\min}.
\]
Conversely, $A_0\mathbf w_0^\star=\bm\beta_0$ and $0\le\mathbf w_0^\star\le K\bm1$, so the primal representation gives
\[
B_K(\eta_0)
\le
a(A_0)+\bm c(A_0)^\top\mathbf w_0^\star
=
\theta_{\min}.
\]
Thus $B_K(\eta_0)=\theta_{\min}$.

It remains to control the plug-in error. This step uses the fact that localization puts the value map on a common bounded feasible set. Let
\[
\mathcal W_K:=\{\mathbf w:0\le \mathbf w\le K\bm1\},
\qquad
\Phi_K(\eta,\mathbf w)
:=
a(A)+\bm c(A)^\top\mathbf w+K\|A\mathbf w-\bm\beta\|_1,
\]
so that $B_K(\eta)=\min_{\mathbf w\in\mathcal W_K}\Phi_K(\eta,\mathbf w)$. For any two parameters $\eta=(A,\bm\beta)$ and $\tilde\eta=(\tilde A,\tilde{\bm\beta})$,
\[
|B_K(\eta)-B_K(\tilde\eta)|
\le
\sup_{\mathbf w\in\mathcal W_K}
|\Phi_K(\eta,\mathbf w)-\Phi_K(\tilde\eta,\mathbf w)|.
\]
We now bound the right-hand side explicitly. Since all norms on the finite-dimensional parameter space are equivalent and $A\mapsto a(A)$ and $A\mapsto\bm c(A)$ are linear, there exist finite constants $C_a,C_c,C_A,C_\beta$ such that, uniformly over $\mathbf w\in\mathcal W_K$,
\[
|a(A)-a(\tilde A)|
\le C_a\|\eta-\tilde\eta\|,
\qquad
|(\bm c(A)-\bm c(\tilde A))^\top\mathbf w|
\le C_cK\|\eta-\tilde\eta\|,
\]
and
\[
\|(A-\tilde A)\mathbf w-(\bm\beta-\tilde{\bm\beta})\|_1
\le
(C_AK+C_\beta)\|\eta-\tilde\eta\|.
\]
By the reverse triangle inequality,
\begin{align*}
|\Phi_K(\eta,\mathbf w)-\Phi_K(\tilde\eta,\mathbf w)|
&\le
|a(A)-a(\tilde A)|
+|(\bm c(A)-\bm c(\tilde A))^\top\mathbf w|\\
&\quad
+K\|(A-\tilde A)\mathbf w-(\bm\beta-\tilde{\bm\beta})\|_1\\
&\le
\{C_a+C_cK+K(C_AK+C_\beta)\}\|\eta-\tilde\eta\|.
\end{align*}
Thus, for this fixed radius $K$, there is a finite constant $L_K$ such that
\[
|B_K(\eta)-B_K(\tilde\eta)|
\le
L_K\|\eta-\tilde\eta\|.
\]
Applying this Lipschitz bound at $\eta=\hat\eta_n$ and $\tilde\eta=\eta_0$ gives
\[
|B_K(\hat\eta_n)-\theta_{\min}|
=
|B_K(\hat\eta_n)-B_K(\eta_0)|
\le
L_K\|\hat\eta_n-\eta_0\|.
\]
Assumption~\ref{assmp:convergence-rate-theta} implies $\|\hat\eta_n-\eta_0\|=O_p(\tau_n^{-1})$, so $B_K(\hat\eta_n)-\theta_{\min}=O_p(\tau_n^{-1})$.

\subsubsection{Proof of Theorem~\ref{thm:consistency-theta}}

We first show that the selected radius is asymptotically equal to the population radius. Let
\[
j_*:=\min\{j:\Gamma_{L_j}(\eta_0)=0\},
\qquad
K_*:=L_{j_*}.
\]
Because the population LP has a finite primal-dual optimal pair and $L_j\to\infty$, Lemma~\ref{lem:Gamma-zero-main} implies that $j_*<\infty$. Since $j_*$ is fixed, it belongs to $\{1,\ldots,n\}$ for all large $n$.

We use a uniform Lipschitz bound for the certification gap. Let
\[
\Psi_K(\eta;\mathbf w,\bm\lambda)
:=
\bm c(A)^\top\mathbf w-\bm\beta^\top\bm\lambda
+2K\|A\mathbf w-\bm\beta\|_1
+2K\bigl\|\bigl[A^\top\bm\lambda-\bm c(A)\bigr]_+\bigr\|_1,
\]
with $(\mathbf w,\bm\lambda)$ restricted to
\[
\mathcal K_K:=\{(\mathbf w,\bm\lambda):0\le \mathbf w\le K\bm1,\ \|\bm\lambda\|_\infty\le K\}.
\]
For any $\eta=(A,\bm\beta)$ and $\tilde\eta=(\tilde A,\tilde{\bm\beta})$, the difference in the criterion satisfies
\begin{align*}
&|\Psi_K(\eta;\mathbf w,\bm\lambda)-\Psi_K(\tilde\eta;\mathbf w,\bm\lambda)|\\
&\quad\le
|(\bm c(A)-\bm c(\tilde A))^\top\mathbf w|
+|(\bm\beta-\tilde{\bm\beta})^\top\bm\lambda|\\
&\qquad
+2K\|(A-\tilde A)\mathbf w-(\bm\beta-\tilde{\bm\beta})\|_1
+2K\|(A-\tilde A)^\top\bm\lambda-(\bm c(A)-\bm c(\tilde A))\|_1.
\end{align*}
Because $\mathbf w$ and $\bm\lambda$ are both bounded by $K$ on $\mathcal K_K$, and because $A\mapsto\bm c(A)$ is linear, finite-dimensional norm equivalence gives a constant $L>0$ such that, uniformly over $\mathcal K_K$ and all $K\ge1$,
\[
|\Psi_K(\eta;\mathbf w,\bm\lambda)-\Psi_K(\tilde\eta;\mathbf w,\bm\lambda)|
\le
LK^2\|\eta-\tilde\eta\|.
\]
Taking infima over the common compact set $\mathcal K_K$ gives
\[
|\Gamma_K(\eta)-\Gamma_K(\tilde\eta)|
\le
LK^2\|\eta-\tilde\eta\|.
\]

Therefore, by Assumption~\ref{assmp:convergence-rate-theta},
\[
\sup_{1\le j\le n}|\Gamma_{L_j}(\hat\eta_n)-\Gamma_{L_j}(\eta_0)|
\le
L L_n^2\|\hat\eta_n-\eta_0\|
=
o_p(\delta_n),
\]
where the last equality uses $L_n^2/\tau_n\to0$ and $\tau_n\delta_n/L_n^2\to\infty$.

For any $j<j_*$, Lemma~\ref{lem:Gamma-zero-main} and the definition of $j_*$ imply $\Gamma_{L_j}(\eta_0)>0$. Since there are finitely many such indices,
\[
\underline\gamma:=\min_{1\le j<j_*}\Gamma_{L_j}(\eta_0)>0
\]
whenever $j_*>1$. Thus no $j<j_*$ passes the sample certification rule with probability tending to one. At the same time, $\Gamma_{K_*}(\eta_0)=0$, so $\Gamma_{K_*}(\hat\eta_n)=o_p(\delta_n)$ and $j_*$ passes the rule with probability tending to one. Therefore $\hat j_n=j_*$ with probability tending to one, and hence
\[
\hat j_n\to_P j_*,
\qquad
\widehat K_n\to_P K_*.
\]

On the event $E_n:=\{\widehat K_n=K_*\}$, which satisfies $\P(E_n)\to1$, we have
\[
\hat\theta_{\min}=B_{K_*}(\hat\eta_n).
\]
By Lemma~\ref{lem:Gamma-zero-main}, $K_*$ contains a population primal-dual optimal pair. The exactness argument in Proposition~\ref{prop:fix-penalty} therefore gives $B_{K_*}(\eta_0)=\theta_{\min}$. The fixed-radius Lipschitz argument also gives a finite constant $L_{K_*}$ such that
\[
|\hat\theta_{\min}-\theta_{\min}|
\le
L_{K_*}\|\hat\eta_n-\eta_0\|
\]
on $E_n$. Assumption~\ref{assmp:convergence-rate-theta} implies the right-hand side is $O_p(\tau_n^{-1})$, and the complement of $E_n$ has probability tending to zero. Hence $B_{\widehat K_n}(\hat\eta_n)-\theta_{\min}=O_p(\tau_n^{-1})$.

\subsubsection{Proof of Proposition~\ref{prop:limit-distribution}}

The selection step is asymptotically inactive. On $E_n=\{\widehat K_n=K_*\}$,
\[
\tau_n(\hat\theta_{\min}-\theta_{\min})
=
\tau_n\{B_{K_*}(\hat\eta_n)-B_{K_*}(\eta_0)\}.
\]
It remains to justify the fixed-radius directional delta method and to make the derivative explicit. Throughout this proof, fix a radius $K$ and write
\[
A_0,\quad \bm\beta_0,\quad \bm c_0:=\bm c(A_0),
\qquad
v_K:=B_K(\eta_0)-a(A_0).
\]
The scalar $v_K$ is the optimal value of the inner localized LP, after removing the constant term $a(A_0)$. Its primal and dual optimal faces are
\[
\mathcal P_K^\star
:=
\left\{
(\mathbf w,\mathbf r^+,\mathbf r^-):
\begin{array}{l}
A_0\mathbf w+\mathbf r^+-\mathbf r^-=\bm\beta_0,\quad
0\le \mathbf w\le K\bm1,\quad \mathbf r^+,\mathbf r^-\ge0,\\[2mm]
\bm c_0^\top\mathbf w+K\bm1^\top(\mathbf r^++\mathbf r^-)=v_K
\end{array}
\right\},
\]
and
\[
\mathcal D_K^\star
:=
\left\{
(\bm\lambda,\mathbf u):
\begin{array}{l}
A_0^\top\bm\lambda-\mathbf u\le\bm c_0,\quad
-K\bm1\le\bm\lambda\le K\bm1,\quad \mathbf u\ge0,\\[1mm]
\bm\beta_0^\top\bm\lambda-K\bm1^\top\mathbf u=v_K
\end{array}
\right\}.
\]
Both sets are nonempty compact polyhedra for the fixed radius used below.

Consider a deterministic direction $h=(\dot A,\dot{\bm\beta})$ and define
\[
\dot a:=\bm1^\top \dot A D\bm1,
\qquad
\dot{\bm c}:=D\dot A^\top\bm1.
\]
The finite-dimensional specialization of the directional value formula in \citet[Theorem~3.5]{shapiro1991asymptotic} gives
\begin{equation}\label{eq:apx-BK-derivative-minmax}
B'_{K,\eta_0}(h)
=
\dot a
+
\min_{(\mathbf w,\mathbf r^+,\mathbf r^-)\in\mathcal P_K^\star}
\max_{(\bm\lambda,\mathbf u)\in\mathcal D_K^\star}
\left\{
\dot{\bm c}^{\top}\mathbf w
+
\bm\lambda^\top(\dot{\bm\beta}-\dot A\mathbf w)
\right\}.
\end{equation}
The term $\dot{\bm c}^{\top}\mathbf w$ is the first-order change in the objective loading, while $\bm\lambda^\top(\dot{\bm\beta}-\dot A\mathbf w)$ is the first-order change in the equality restriction $A\mathbf w=\bm\beta$ evaluated at a dual multiplier.

For completeness, and also for computation, we now write \eqref{eq:apx-BK-derivative-minmax} as a single LP. For a fixed primal optimizer $\mathbf w$, the inner maximization in \eqref{eq:apx-BK-derivative-minmax} is the LP
\[
\max_{(\bm\lambda,\mathbf u)\in\mathcal D_K^\star}
\bm\lambda^\top(\dot{\bm\beta}-\dot A\mathbf w).
\]
Dualizing this inner LP introduces variables
\[
\bm\zeta\in\mathbb R_+^M,\qquad
\bm\xi^+,\bm\xi^-\in\mathbb R_+^{|\mathcal F|},\qquad
\bm\omega\in\mathbb R_+^M,\qquad
\tau\in\mathbb R,
\]
corresponding respectively to the constraints
$A_0^\top\bm\lambda-\mathbf u\le\bm c_0$,
$\bm\lambda\le K\bm1$,
$-\bm\lambda\le K\bm1$,
$-\mathbf u\le0$, and
$\bm\beta_0^\top\bm\lambda-K\bm1^\top\mathbf u=v_K$. Hence, for every direction $h=(\dot A,\dot{\bm\beta})$,
\begin{equation}\label{eq:apx-BK-derivative-lp}
\begin{aligned}
B'_{K,\eta_0}(h)
=\dot a+
\min_{\substack{
\mathbf w,\mathbf r^+,\mathbf r^-,\\
\bm\zeta,\bm\xi^+,\bm\xi^-,\bm\omega,\tau}}
&\quad
\dot{\bm c}^{\top}\mathbf w
+\bm c_0^\top\bm\zeta
+K\bm1^\top(\bm\xi^++\bm\xi^-)
+v_K\tau\\
\text{s.t.}
&\quad
A_0\mathbf w+\mathbf r^+-\mathbf r^-=\bm\beta_0,\\
&\quad
0\le \mathbf w\le K\bm1,\qquad
\mathbf r^+,\mathbf r^-\ge0,\\
&\quad
\bm c_0^\top\mathbf w+K\bm1^\top(\mathbf r^++\mathbf r^-)=v_K,\\
&\quad
A_0\bm\zeta+\bm\xi^+-\bm\xi^-+\tau\bm\beta_0
=
\dot{\bm\beta}-\dot A\mathbf w,\\
&\quad
\bm\zeta+\bm\omega+\tau K\bm1=\bm0,\\
&\quad
\bm\zeta,\bm\xi^+,\bm\xi^-,\bm\omega\ge0,\qquad
\tau\in\mathbb R .
\end{aligned}
\end{equation}
This is a linear program because the direction $(\dot A,\dot{\bm\beta})$ is fixed, so $\dot A\mathbf w$ is linear in the decision variable $\mathbf w$.

The same value formula also gives Hadamard directional differentiability of the fixed-radius value map. To see this directly in our notation, fix a compact neighborhood $\mathcal U$ of $\eta_0$ and define
\[
\bar s
:=
2+2\sup_{\eta\in\mathcal U}\sup_{\|\bm\lambda\|_\infty\le K_*}
\bigl\|A^\top\bm\lambda-\bm c(A)\bigr\|_\infty.
\]
Then $\bar s<\infty$. Using \eqref{eq:apx-BK-dual}, for every $\eta\in\mathcal U$,
\[
B_{K_*}(\eta)
=
a(A)
+
\max_{(\bm\lambda,\mathbf s)\in\widetilde{\mathcal D}_{K_*}}
\Bigl\{\bm\beta^\top\bm\lambda-K_*\bm1^\top\mathbf s:
A^\top\bm\lambda-\bm c(A)-\mathbf s\le0\Bigr\},
\]
where
\[
\widetilde{\mathcal D}_{K_*}:=\{(\bm\lambda,\mathbf s):\|\bm\lambda\|_\infty\le K_*,\ 0\le\mathbf s\le\bar s\bm1\}
\]
is a fixed compact set. The optimizer in $\mathbf s$ is $[A^\top\bm\lambda-\bm c(A)]_+$, which belongs to this set by construction. The point $(0,(\bar s/2)\bm1)$ satisfies the inequality constraints strictly throughout a sufficiently small neighborhood of $\eta_0$, so a local Slater condition holds. Because the objective and constraint maps are affine in $\eta$, standard value-function results for compact convex programs imply that $\eta\mapsto B_{K_*}(\eta)$ is Hadamard directionally differentiable at $\eta_0$, with derivative given by \eqref{eq:apx-BK-derivative-lp} evaluated at $K=K_*$.

Assumption~\ref{assmp:convergence-rate-theta} and the directional delta method therefore imply
\[
\tau_n\{B_{K_*}(\hat\eta_n)-B_{K_*}(\eta_0)\}
\rightsquigarrow
B'_{K_*,\eta_0}(G).
\]
Since $\P(E_n)\to1$, the same limit holds for the selected-radius estimator. This is the stated limit with $B'_{K_*}(\cdot)=B'_{K_*,\eta_0}(\cdot)$.

\subsubsection{Proof of Theorem~\ref{thm:subsample-bootstrap}}

Let
\[
T_n:=\tau_n(\hat\theta_{\min}-\theta_{\min}),
\]
and let $\mathcal L$ denote its weak limit. By Proposition~\ref{prop:limit-distribution}, $T_n\rightsquigarrow\mathcal L$. On the event $\{\widehat K_n=K_*\}$, the subsampling statistic can be written as
\[
T_{m,S}
=
\tau_m\{B_{K_*}(\hat\eta_{m,S})-B_{K_*}(\eta_0)\}
-
\frac{\tau_m}{\tau_n}\,
\tau_n\{B_{K_*}(\hat\eta_n)-B_{K_*}(\eta_0)\}.
\]
The second term is $o_p(1)$ because $\tau_m/\tau_n\to0$ and the full-sample root is tight. Thus the subsampling distribution is asymptotically equivalent to the usual fixed-radius subsampling distribution for the statistic $\tau_n\{B_{K_*}(\hat\eta_n)-B_{K_*}(\eta_0)\}$.

As $m\to\infty$, the same argument as in Proposition~\ref{prop:limit-distribution} applied to an i.i.d.\ sample of size $m$ gives
\[
\tau_m\{B_{K_*}(\hat\eta_{m})-B_{K_*}(\eta_0)\}\rightsquigarrow\mathcal L.
\]
Together with the full-sample convergence $T_n\rightsquigarrow\mathcal L$, this verifies the input condition of the large-sample subsampling theorem of \citet[Theorem~3.1]{politis1994large}. Applying that result to the fixed-radius statistic gives
\[
\hat J_{n,m}(t)\to_P J_{\mathcal L}(t)
\]
at every continuity point $t$ of $J_{\mathcal L}$ whenever $m\to\infty$ and $m/n\to0$. Quantile consistency in part (b) follows from the usual inversion argument under the stated separation condition. Finally, under the separation and continuity conditions in part (c), Slutsky's theorem and \eqref{eq:fs-endpoint-ci} imply the coverage statement.

\subsubsection{Proof of Proposition~\ref{prop:ci-identification-region}}

Define
\[
\underline C_{1-\alpha}
:=
\hat\theta_{\min,n}-\tau_n^{-1}\hat q_{n,m}^{L}(1-\alpha/2),
\qquad
\overline C_{1-\alpha}
:=
\hat\theta_{\max,n}-\tau_n^{-1}\hat q_{n,m}^{U}(\alpha/2).
\]
By Theorem~\ref{thm:subsample-bootstrap} applied to the lower endpoint,
\[
\liminf_{n\to\infty}\P(\theta_{\min}\ge\underline C_{1-\alpha})\ge1-\alpha/2.
\]
The same argument for the upper endpoint gives
\[
\liminf_{n\to\infty}\P(\theta_{\max}\le\overline C_{1-\alpha})\ge1-\alpha/2.
\]
Therefore, by Bonferroni's inequality,
\begin{align*}
\liminf_{n\to\infty}
\P\bigl(\Theta_0\subseteq\widehat{\mathrm{CR}}_{1-\alpha}\bigr)
&=
\liminf_{n\to\infty}
\P(\theta_{\min}\ge\underline C_{1-\alpha},\ \theta_{\max}\le\overline C_{1-\alpha})\\
&\ge
1-\alpha.
\end{align*}

\subsubsection{Proof of Proposition~\ref{prop:continuous-x-cmom}}

The proposition is the conditional-moment version of the same accounting identity used in the pointwise LP. For each $\ell\in\mathcal F$ and almost every $x$,
\[
\E[\mathbf 1\{R=0,F=\ell\}\mid X=x]=\beta_x(\ell),
\]
and
\[
\E[\mathbf 1\{R=1,F=\ell,RY=y\}w(y,X)\mid X=x]
=
\alpha_x(\ell,y)w_x(y).
\]
Therefore,
\[
\E[m_\ell(O;w)\mid X=x]
=
\beta_x(\ell)-\sum_{y=1}^M\alpha_x(\ell,y)w_x(y).
\]
This conditional expectation is zero for every $\ell$ if and only if the pointwise feasibility restriction
$\sum_{y=1}^M\alpha_x(\ell,y)w_x(y)=\beta_x(\ell)$ holds for $P_X$-almost every $x$.

If \eqref{eq:cont-x-cmom} holds, then multiplying by any integrable test function $q(X)$ and taking expectations gives \eqref{eq:cont-x-ucmom}. Conversely, suppose \eqref{eq:cont-x-ucmom} holds for a class $\mathcal Q$ that characterizes conditional mean zero. Applying this property to $g_\ell(X):=\E[m_\ell(O;w)\mid X]$ gives $g_\ell(X)=0$ almost surely for every $\ell\in\mathcal F$. Hence the conditional moment restrictions, and therefore the pointwise feasibility restrictions, hold.

\end{APPENDICES}

\end{document}